\RequirePackage{fix-cm}
\documentclass[smallcondensed,natbib]{svjour3}

\usepackage{amsmath,amssymb}
\usepackage{booktabs}
\usepackage{float}
\usepackage{graphicx}
\usepackage{longtable}
\usepackage[caption=false]{subfig}
\usepackage[colorlinks=true,linkcolor=blue,urlcolor=blue,citecolor=blue]{hyperref}

\begin{document}

\title{
  ROCKET: Exceptionally fast and accurate time series classification using random convolutional kernels
}
\titlerunning{
  \textsc{ROCKET: Exceptionally fast and accurate time series classification}
}

\author{
  Angus~Dempster         \and
  Fran\c{c}ois~Petitjean \and
  Geoffrey~I.~Webb
}

\institute{
  Angus Dempster \and Fran\c{c}ois Petitjean \and Geoffrey I. Webb
  \at
  Faculty of Information Technology, Monash University, Melbourne, Australia \\
  \email{\{angus.dempster1,francois.petitjean,geoff.webb\}@monash.edu}
}

\date{Received: date / Accepted: date}

\maketitle

\begin{abstract}
Most methods for time series classification that attain state-of-the-art accuracy have high computational complexity, requiring significant training time even for smaller datasets, and are intractable for larger datasets.  Additionally, many existing methods focus on a single type of feature such as shape or frequency.  Building on the recent success of convolutional neural networks for time series classification, we show that simple linear classifiers using random convolutional kernels achieve state-of-the-art accuracy with a fraction of the computational expense of existing methods.
\end{abstract}

\section{Introduction}

\begin{figure}
\centering
\vspace*{-4mm}
\includegraphics[width=0.9\linewidth]{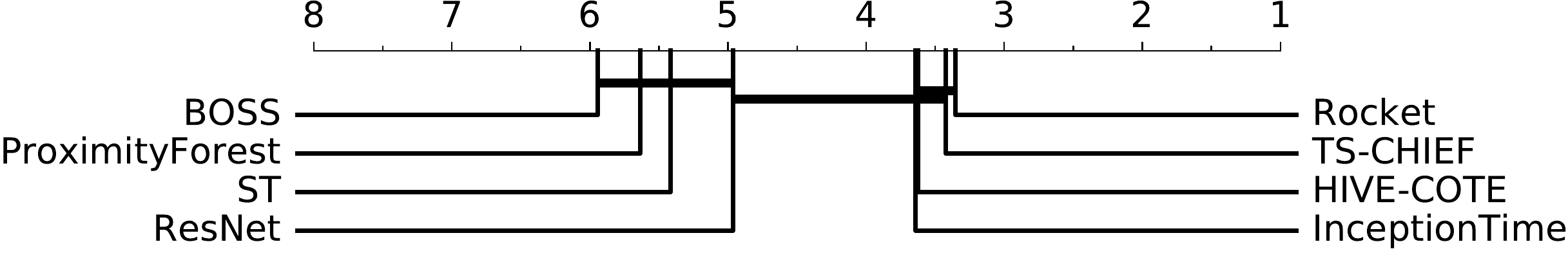}
\vspace*{-2mm}
\caption{Mean rank of \textsc{Rocket} versus state-of-the-art classifiers on the 85 `bake off' datasets.}
\label{figure-rank-ucr85}
\end{figure}

Most methods for time series classification that attain state-of-the-art accuracy have high computational complexity, requiring significant training time even for smaller datasets, and simply do not scale to large datasets.  This has motivated the development of more scalable methods such as Proximity Forest \citep{lucas_etal_2019}, TS-CHIEF\,\citep{shifaz_etal_2019}, and InceptionTime\,\citep{ismailfawaz_etal_2019_c}.

We show that state-of-the-art classification accuracy can be achieved using a fraction of the time required by even these recent, more scalable methods, by transforming time series using random convolutional kernels, and using the transformed features to train a linear classifier.  We call this method \textsc{Rocket} (for \textbf{R}and\textbf{O}m \textbf{C}onvolutional \textbf{KE}rnel \textbf{T}ransform).

Existing methods for time series classification typically focus on a single representation such as shape, frequency, or variance.  Convolutional kernels constitute a single mechanism which can capture many of the features which have each previously required their own specialized techniques, and have been shown to be effective in convolutional neural networks for time series classification such as ResNet \citep{wang_etal_2017,ismailfawaz_etal_2019_a}, and InceptionTime.

In contrast to learned convolutional kernels as used in typical convolutional neural networks, we show that it is effective to generate a large number of random convolutional kernels which, in combination, capture features relevant for time series classification (even though, in isolation, a single random convolutional kernel may only very approximately capture a relevant feature in a given time series).

\textsc{Rocket} achieves state-of-the-art classification accuracy on the datasets in the UCR archive \citep{dau_etal_2019}, but requires only a fraction of the training time of existing methods.  Figure~\ref{figure-rank-ucr85} shows the mean rank of \textsc{Rocket} versus several state-of-the-art methods for time series classification on the 85 `bake off' datasets from the UCR archive \citep{dau_etal_2019,bagnall_etal_2017}.  Restricted to a single CPU core, the total training time for \textsc{Rocket} is:

\begin{itemize}
  \item 6 minutes for the `bake off' dataset with the largest training set (\textit{ElectricDevices}, with 8,926 training examples), compared to 1 hour 35 minutes for Proximity Forest, 2 hours 24 minutes for TS-CHIEF, and 7 hours 46 minutes for InceptionTime (trained on GPUs); and
  \item 4 minutes and 52 seconds for the `bake off' dataset with the longest time series (\textit{HandOutlines}, with time series of length 2,709), compared to 8 hours 10 minutes for InceptionTime (trained on GPUs), almost 3 days for Proximity Forest, and more than 4 days for TS-CHIEF.
\end{itemize}

The total compute time (training and test) for \textsc{Rocket} on all 85 `bake off' datasets is 1 hour 50 minutes, compared to more than 6 days for InceptionTime (trained and tested using GPUs), and more than 11 days for each of Proximity Forest and TS-CHIEF.  (Timings for \textsc{Rocket} are averages over 10 runs, performed on a cluster using a mixture of Intel Xeon E5-2680 v3 and Intel Xeon Gold 6150 processors, restricted to a single CPU core per dataset per run.)

\textsc{Rocket} is also more scalable for large datasets, with training complexity linear in both time series length and the number of training examples.  \textsc{Rocket} can learn from 1 million time series in 1 hour 15 minutes, to a similar accuracy as Proximity Forest, which requires more than 16 hours to train on the same quantity of data.  A restricted variant of \textsc{Rocket} can learn from the same 1 million time series in less than 1 minute, or approximately 100 times faster again, albeit to a slightly lower accuracy.  \textsc{Rocket} is naturally parallel, and can be made even faster by using multiple CPU cores (our implementation automatically parallelises the transform across multiple CPU cores where available) or GPUs.

The rest of this paper is structured as follows.  In section \ref{section-related-work}, we review relevant related work.  In section \ref{section-method}, we explain \textsc{Rocket} in detail.  In section \ref{section-experiments}, we present our experimental results, including a comparison of the accuracy of \textsc{Rocket} against existing state-of-the-art classifiers on the datasets in the UCR archive, a scalability study, and a sensitivity analysis.

\section{Related Work} \label{section-related-work}

\subsection{State-of-the-Art Methods}

The task of time series classification can be thought of as involving learning or detecting signals or patterns within time series associated with relevant classes.  `[D]ifferent problems require different representations' \citep[p. 647]{bagnall_etal_2017}, and classes may be distinguished by multiple types of patterns: `discriminatory features in multiple domains' \citep[p. 645]{bagnall_etal_2017}.

Different methods for time series classification represent different approaches for extracting useful features from time series \citep{bagnall_etal_2017}.  Existing approaches typically focus on a single type of feature, such as frequency or variance of the signal, or the presence of discriminative subseries (shapelets).  \citet{bagnall_etal_2017} identified COTE (since superseded by HIVE-COTE), Shapelet Transform \citep{hills_etal_2014,bostrom_and_bagnall_2015}, and BOSS \citep{schafer_2015} as the three most accurate classifiers on the UCR archive.

BOSS is one of several dictionary-based methods which use a representation based on the frequency of occurrence of patterns in time series \citep{bagnall_etal_2017}.  BOSS has a training complexity quadratic in both the number of training examples and time series length, $O(n^{2} \cdot l^{2})$.  BOSS-VS is a more scalable variant of BOSS, but is less accurate \citep{schafer_2016}.  Another related method, WEASEL, is more accurate than BOSS, but with a similar training complexity and high memory complexity (\citealt{schafer_and_leser_2017}; see also \citealt{lucas_etal_2019}).

Shapelet Transform is one of several methods based on finding discriminative subseries, so-called `shapelets' \citep{bagnall_etal_2017}.  Shapelet Transform has a training complexity quadratic in the number of training examples, and quartic in time series length, $O(n^{2} \cdot l^{4})$.  There are other, more scalable, shapelet methods, but these are less accurate \citep{bagnall_etal_2017}.

HIVE-COTE is a large ensemble of other classifiers, including BOSS and Shapelet Transform, as well as classifiers based on elastic distance measures and frequency representations \citep{lines_etal_2018}.  Since \citet{lines_etal_2018}, HIVE-COTE has been considered the most accurate method for time series classification.  The training complexity of HIVE-COTE is bound by the complexity of Shapelet Transform, $O(n^{2} \cdot l^{4})$, but its other components also have high computational complexity, such as the Elastic Ensemble with $O(n^{2} \cdot l^{2})$ \citep{lines_etal_2018}.

\subsection{More Scalable Methods}

The high computational complexity of existing state-of-the-art methods for time series classification makes these methods slow, even for smaller datasets, and intractable for large datasets.  This has motivated the development of more scalable methods, including Proximity Forest, TS-CHIEF, and InceptionTime.

Proximity Forest is an ensemble of decision trees, using elastic distance measures as splitting criteria \citep{lucas_etal_2019}, with a training complexity quasilinear in the number of training examples, but quadratic in time series length.

TS-CHIEF builds on Proximity Forest, incorporating dictionary-based and interval-based splitting criteria \citep{shifaz_etal_2019}.  Like Proximity Forest, TS-CHIEF has a training complexity quasilinear in the number of training examples, but quadratic in time series length.

Several methods for time series classification using convolutional neural networks have been proposed \citep[see generally][]{ismailfawaz_etal_2019_a}.  More recently, InceptionTime \citep{ismailfawaz_etal_2019_c}, an ensemble of five deep convolutional neural networks based on the Inception architecture, has been demonstrated to be competitive with HIVE-COTE on the UCR archive.

Convolutional neural networks are typically trained using stochastic gradient descent or closely related algorithms such as, for example, Adam \citep{kingma_and_ba_2015}.  The training complexity of stochastic gradient descent is essentially linear with respect to the number of training examples, and training can be parallelised using GPUs (\citealt[pp. 147--149]{goodfellow_etal_2016}; \citealt{bottou_etal_2018}).

\subsection{Convolutional Neural Networks and Convolutional Kernels}

\citet[pp. 2--3]{ismailfawaz_etal_2019_c} observe that the success of convolutional neural networks for image classification suggests that they should also be effective for time series classification, given that time series have essentially the same topology as images, with one less dimension \citep[see also][p. 1820--1821]{bengio_etal_2013}.

Convolutional neural networks represent a different approach to time series classification than many other methods.  Rather than approaching the problem with a preconceived representation, convolutional neural networks use convolutional kernels to detect patterns in the input.  In learning the weights of the kernels, a convolutional neural network learns the features in time series associated with different classes \citep{ismailfawaz_etal_2019_a}.

A kernel is convolved with an input time series through a sliding dot product operation, to produce a feature map which is, in turn, used as the basis for classification \citep[see][]{ismailfawaz_etal_2019_a}.  The basic parameters of a kernel are its size (length), weights and bias, dilation, and padding \citep[see generally][ch. 9]{goodfellow_etal_2016}.  A kernel has the same structure as the input, but is typically much smaller.  For time series, a kernel is a vector of weights, with a bias term which is added to the result of the convolution operation between an input time series and the weights of the given kernel.  Dilation `spreads' a kernel over the input such that with a dilation of two, for example, the weights in a kernel are convolved with every second element of an input time series \citep[see][]{bai_etal_2018}.  Padding involves appending values (typically zero) to the start and end of input time series, typically such that the `middle' weight of a given kernel aligns with the first element of an input time series at the start of the convolution operation.

Convolutional kernels can capture many of the types of features used in other methods.  Kernels can capture basic patterns or shapes in time series, similar to shapelets: the convolution operation will produce large output values where the kernel matches the input.  Further, dilation allows kernels to capture the same pattern at different scales \citep{yu_and_koltun_2016}.  Multiple kernels in combination can capture complex patterns.

The feature maps produced in applying a kernel to a time series reflect the extent to which the pattern represented by the kernel is present in the time series.  In a sense, this is not unlike dictionary methods, which are based on the frequency of occurrence of patterns in time series.

The kernels learned in convolutional neural networks often include filters for frequency \citep[see, e.g.,][]{krizhevsky_etal_2012,yosinski_etal_2014,zeiler_and_fergus_2014}.  \citet{saxe_etal_2011} demonstrate that even random kernels are frequency selective.  Frequency information is also captured through dilation: larger dilations correspond to lower frequencies, smaller dilations to higher frequencies.

Kernels can detect patterns in time series despite warping.  Pooling mechanisms make kernels invariant to the position of patterns in time series.  Dilation allows kernels with similar weights to capture patterns at different scales, i.e., despite rescaling.  Multiple kernels with different dilations can, in combination, capture discriminative patterns despite complex warping.

The success of convolutional neural networks for time series classification, such as ResNet and InceptionTime, demonstrates the effectiveness of convolutional kernels as the basis for time series classification.

\subsection{Random Convolutional Kernels}

The weights of convolutional kernels are typically learned.  However, it is well established that random convolutional kernels can be effective \citep{jarrett_etal_2009,pinto_etal_2009,saxe_etal_2011,cox_and_pinto_2011}.

\citet{ismailfawaz_etal_2019_c} observe that individual convolutional neural networks exhibit high variance in classification accuracy on the UCR archive, motivating the use of ensembles of such architectures with a large number and variety of kernels \citep[see][]{ismailfawaz_etal_2019_b}.  It may be that learning `good' kernels is difficult on small datasets.  Random convolutional kernels may have an advantage in this context \citep[see][]{jarrett_etal_2009,yosinski_etal_2014}.

The idea of using convolutional kernels as a transform, and using the transformed features as the input to another classifier is well established \citep[see, e.g.,][1803]{bengio_etal_2013}.  \citet{franceschi_etal_2019} present a method for unsupervised learning of convolutional kernels for a feature transform for time series input, based on a multilayer convolutional architecture with dilation increasing exponentially in each successive layer.  The method is demonstrated using the output features as the input for a support vector machine.

Random convolutional kernels have been used as the basis of feature transformations.  In \citet{saxe_etal_2011}, random convolutional layers are used as the basis of a feature transform (for images), used as the input for a support vector machine.

Here, there is a link between using random convolutional kernels as a transform for time series and work in relation to random transforms for kernel methods (as in support vector machines, not to be confused with convolutional kernels).  \citet{rahimi_and_recht_2008} proposed a random transform for approximating kernels for kernel methods \citep[see also][]{rahimi_and_recht_2009}.  \citet{morrow_etal_2017} propose a method for approximating a string kernel for DNA sequences, based on \citet{rahimi_and_recht_2008}, which involves transforming input sequences using random convolutional kernels, and using the resulting features to train a linear classifier.  \citet[p. 1]{morrow_etal_2017} describe their method as `a 1 layer random convolutional neural network'.  Also following \citet{rahimi_and_recht_2008}, \citet{jimenez_and_raj_2019} propose a similar method for approximating a cross-correlation kernel for measuring similarity between time series, involving convolving input time series with random time series of the same length to produce what they call `random convolutional features', which can be used to train a linear classifier.  \citet{jimenez_and_raj_2019} evaluate their method on a selection of binary classification datasets from the UCR archive.  (In both cases, there are some differences with the convolution operation as used in typical convolutional neural networks.)  \citet{farahmand_etal_2017} propose a feature transformation based on convolving input time series with random autoregressive filters.

A number of things distinguish \textsc{Rocket} from convolutional layers as used in typical convolutional neural networks, and from other methods using convolutional kernels (including random convolutional kernels) in relation to time series, set out in detail in section \ref{section-method}.  We show that leveraging all aspects of kernel architecture---crucially, with a variety of random length, dilation, and padding (as well as weights and bias), and drawing an effective set of features from the output of the convolutions---provides for state-of-the-art accuracy with a fraction of the computational expense of existing state-of-the-art methods.

\section{Method} \label{section-method}

\textsc{Rocket} transforms time series using a large number of random convolutional kernels, i.e., kernels with random length, weights, bias, dilation, and padding.  The transformed features are used to train a linear classifier.  The combination of \textsc{Rocket} and logistic regression forms, in effect, a single-layer convolutional neural network with random kernel weights, where the transformed features form the input for a trained softmax layer.  However, in practice, for all but the largest datasets, we use a ridge regression classifier, which has the advantage of fast cross-validation for the regularization hyperparameter (and no other hyperparameters).  Nonetheless, as logistic regression trained using stochastic gradient descent is more scalable for very large datasets, we use logistic regression when the number of training examples is substantially greater than the number of features.

Four things distinguish \textsc{Rocket} from convolutional layers as used in typical convolutional neural networks, and from previous work using convolutional kernels (including random kernels) with time series:

\begin{enumerate}
  \item \textsc{Rocket} uses a very large number of kernels.  As there is only a single `layer' of kernels, and as the kernel weights are not learned, the computational cost of computing the convolutions is low, and it is possible to use a very large number of kernels with relatively little computational expense.
  \item \textsc{Rocket} uses a massive variety of kernels.  In contrast to typical convolutional networks, where it is common for groups of kernels to share the same size, dilation, and padding, for \textsc{Rocket} each kernel has random length, dilation, and padding, as well as random weights and bias.
  \item In particular, \textsc{Rocket} makes key use of kernel dilation.  In contrast to the typical use of dilation in convolutional neural networks, where dilation increases exponentially with depth \citep[e.g.,][]{yu_and_koltun_2016,bai_etal_2018,franceschi_etal_2019}, we sample dilation randomly for each kernel, producing a huge variety of kernel dilation, capturing patterns at different frequencies and scales, which is critical to the performance of the method (see section \ref{subsubsection-dilation}, below).
  \item As well as using the maximum value of the resulting feature maps (broadly speaking, similar to global max pooling), \textsc{Rocket} uses an additional and, to our knowledge, novel feature: the proportion of positive values (or \textit{ppv}).  This enables a classifier to weight the prevalence of a given pattern within a time series.  This is the single element of the \textsc{Rocket} architecture that is most critical to its outstanding accuracy (see section \ref{subsubsection-features}).
\end{enumerate}

In effect, the only hyperparameter for \textsc{Rocket} is the number of kernels, $k$.  In setting $k$, there is a tradeoff between classification accuracy and computation time.  Generally speaking, a larger value of $k$ results in higher classification accuracy (see section \ref{subsubsection-number-of-kernels}), but at the expense of proportionally longer computation.  (The complexity of the transform is linear with respect to $k$.)  However, even with a very large number of kernels (we use 10,000 by default), \textsc{Rocket} is extremely fast.

We implement \textsc{Rocket} in Python, using just-in-time compilation via Numba \citep{lam_etal_2015}.  For the experiments on the datasets in the UCR archive, we use a ridge regression classifier from scikit-learn \citep{pedregosa_etal_2011}.  For the experiments studying scalability, we integrate \textsc{Rocket} with logistic regression and Adam, implemented using PyTorch \citep{paszke_etal_2017}.  Our code will be made available at \url{https://github.com/angus924/rocket}.

In developing \textsc{Rocket}, we have endeavoured to not overfit the entire UCR archive \citep[see][p. 608]{bagnall_etal_2017}.  At the same time, in order to develop the method, we required representative time series datasets.  Accordingly, we chose to develop the method on a subset of 40 randomly-selected datasets from the 85 `bake off' datasets.  We refer to these as the `development' datasets.  We provide a separate evaluation of the performance of \textsc{Rocket} on the `development' datasets and the remaining `holdout' datasets in Appendix \ref{section-appendix-rank-development-holdout}.

\subsection{Kernels}

\textsc{Rocket} transforms time series using convolutional kernels, as found in typical convolutional neural networks.  Essentially all aspects of the kernels are random: length, weights, bias, dilation, and padding.  For each kernel, these values are set as follows (as determined by experimentation to produce the highest classification accuracy on the `development' datasets):

\begin{itemize}
  \item \textbf{Length}.  Length is selected randomly from $\{7, 9, 11\}$ with equal probability, making kernels considerably shorter than input time series in most cases.
  \item \textbf{Weights}.  The weights are sampled from a normal distribution, $\forall w \in \boldsymbol{W}$, $w~\sim~\mathcal{N}(0, 1)$, and are mean centered after being set, $\omega = \boldsymbol{W} - \overline{\boldsymbol{W}}$.  As such, most weights are relatively small, but can take on larger magnitudes.
  \item \textbf{Bias}.  Bias is sampled from a uniform distribution, $b \sim \mathcal{U}(-1, 1)$.  Only positive values in the feature maps are used (see section \ref{subsection-transform}).  Bias therefore has the effect that two otherwise similar kernels, but with different biases, can `highlight' different aspects of the resulting feature maps by shifting the values in a feature map above or below zero by a fixed amount.
  \item \textbf{Dilation}.  Dilation is sampled on an exponential scale $d = \lfloor 2^{x} \rfloor, x \sim \mathcal{U}(0, A)$, where $A = log_{2} \frac{l_{\text{input}} - 1}{l_{\text{kernel}} - 1}$, which ensures that the effective length of the kernel, including dilation, is up to the length of the input time series, $l_{\text{input}}$.  Dilation allows otherwise similar kernels but with different dilations to match the same or similar patterns at different frequencies and scales.
  \item \textbf{Padding}.  When each kernel is generated, a decision is made (at random, with equal probability) whether or not padding will be used when applying the kernel.  If padding is used, an amount of zero padding is appended to the start and end of each time series when applying the kernel, such that the `middle' element of the kernel is centered on every point in the time series, i.e., $((l_{\text{kernel}} - 1) \times d) / 2$.  Without padding, kernels are not centered at the first and last $\lfloor l_{\text{kernel}} / 2 \rfloor$ points of the time series, and `focus' on patterns in the central regions of time series whereas with padding, kernels also match patterns at the start or end of time series (see also section \ref{subsubsection-transform}).
\end{itemize}

Stride is always one.  We do not apply a nonlinearity such as ReLU to the resulting feature maps (indeed both \textit{ppv} and max are agnostic to ReLU).  Note that the parameters for the weights and bias have been set based on the assumption that, as is standard practice, input time series have been normalized to have a mean of zero and a standard deviation of one \citep[see generally][]{dau_etal_2019}.

As noted above, these parameters were determined to produce the highest classification accuracy on the `development' datasets.  However, as demonstrated in section \ref{subsection-sensitivity-analysis}, below, there are several alternative configurations which produce similar classification accuracy.  Overall, this suggests that our method is likely to generalise well to new problems, and that the kernel parameters are relatively `uninformative' in the Bayesian sense of the word.

\subsection{Transform} \label{subsection-transform}

Each kernel is applied to each input time series, producing a feature map.  The convolution operation involves a sliding dot product between a kernel and an input time series.  The result of applying a kernel, $\omega$, with dilation, $d$, to a given time series, $X$, from position $i$ in $X$, is given by \citep[see, e.g.,][]{bai_etal_2018}:

$$X_{i} * \omega = \sum_{j=0}^{l_{\text{kernel}}-1} X_{i + (j \times d)} \times \omega_{j}.$$

\textsc{Rocket} computes two aggregate features from each feature map, producing two real-valued numbers as features per kernel, and composing our transform:

\begin{itemize}
\item the maximum value (broadly speaking, equivalent to global max pooling); and
\item the proportion of positive values (or \textit{ppv}).
\end{itemize}

Pooling, including global average pooling \citep{lin_etal_2014}, and global max pooling \citep{oquab_etal_2015}, is used in convolutional neural networks for dimensionality reduction and spatial (or temporal) invariance \citep{boureau_etal_2010}.

The other feature computed by \textsc{Rocket} on each feature map is \textit{ppv}.  The \textit{ppv} directly captures the proportion of the input which matches a given pattern.  We found that \textit{ppv} produces meaningfully higher classification accuracy than other features, including the mean (broadly equivalent to global average pooling).

For $k$ kernels, \textsc{Rocket} produces $2k$ features per time series (i.e., \textit{ppv} and max).  For 10,000 kernels (the default), \textsc{Rocket} produces 20,000 features.  For smaller datasets (in fact, for all the datasets in the UCR archive), the number of features is therefore possibly much larger than either the number of examples in the dataset or the number of elements in each time series.

Nevertheless, we find that the features produced by \textsc{Rocket} provide for high classification accuracy when used as the input for a linear classifier, even for datasets where the number of features dwarfs both the number of examples and the length of the time series.

\subsection{Classifier}

The transformed features are used to train a linear classifier.  \textsc{Rocket} can, in principle, be used with any classifier.  We have found that \textsc{Rocket} is very effective when used in conjunction with linear classifiers (which have the capacity to make use of a small amount of information from each of a large number of features).

\paragraph{Logistic regression.}

\begin{sloppypar}
\textsc{Rocket} can be used with logistic regression and stochastic gradient descent.  This is particularly suitable for very large datasets because it provides for fast training with a fixed memory cost (fixed by the size of each minibatch).  The transform can be performed on each minibatch, or on larger tranches of the dataset which are then divided further into minibatches for training.
\end{sloppypar}

\paragraph{Ridge regression.}

However, for all of the datasets in the UCR archive we use a ridge regression classifier.  (A ridge regression model is trained for each class in a `one versus rest' fashion, with $L_{2}$ regularization.)

Regularization is critically important where the number of features is significantly greater than the number of training examples, allowing for the optimization of linear models, and preventing pathological behaviour in iterative optimisation, e.g., for logistic regression \citep[see][pp. 232--233]{goodfellow_etal_2016}.  The ridge regression classifier can exploit generalised cross-validation to determine an appropriate regularization parameter quickly \citep[see][]{rifkin_and_lippert_2007}.  We find that for smaller datasets, a ridge regression classifier is significantly faster in practice than logistic regression, while still achieving high classification accuracy.

\subsection{Complexity Analysis} \label{subsection-complexity-analysis}

The computational complexity of \textsc{Rocket} has two aspects: (1) the complexity of the transform itself; and (2) the complexity of the linear classifier trained using the transformed features.

\subsubsection{Transform} \label{subsubsection-transform}

The transform itself is linear in relation to both: (a) the number of examples; and (b) the length of the time series in a given dataset.  Formally, the computational complexity of the transform is $O(k \cdot n \cdot l_{\text{input}})$, where $k$ is the number of kernels, $n$ is the number of examples, and $l_{input}$ is the length of the time series.  The transform must be applied to both training and test sets.

The convolution operation can be implemented in more than one way, including as a matrix multiplication typical of implementations for convolutional neural networks, and using the fast Fourier transform \citep[see][ch. 9]{goodfellow_etal_2016}.  We implement \textsc{Rocket} simply, `sliding' each kernel along each time series and computing the dot product at each location.  This involves repeated elementwise multiplication and summation, the complexity of which is dictated by the length of the time series, and the length of the kernels (that is, the number of weights in the kernels).  The length of the kernels for \textsc{Rocket} is limited to, at most, 11.  Accordingly, kernel length is a constant factor for the purpose of this analysis.

Dilation increases the effective size of a kernel.  Accordingly, where no padding is used, dilation reduces computational complexity.  Without padding, the convolution is computed with the first element of the kernel starting at the first element of the time series, and ends once the last element of the kernel reaches the last element of the time series.  In an extreme case, for the largest values of dilation, the kernel will `fill' the entire time series, and the number of computations will be the number of weights in the kernel.  However, padding is applied randomly with equal probability, so the reduction in complexity is a constant factor.

Where padding is used, dilation has no effect on complexity: the same number of computations are required regardless of dilation or the effective size of the kernel.  Regardless of dilation, the kernel is centered on the first element of the time series, and `slides' the same number of elements along the time series.

Accordingly, for $k$ kernels and $n$ time series, each of length $l_{\text{input}}$, the complexity of the transform is $O(k \cdot n \cdot l_{\text{input}})$.  For datasets with time series of different lengths, this could be taken to represent average complexity for an average length of $l_{\text{input}}$, or worst-case complexity for a maximum length of $l_{\text{input}}$.

\subsubsection{Classifier}

\paragraph{Logistic regression and stochastic gradient descent.}

The complexity of stochastic gradient descent is proportional to the number of parameters (dictated by the number of features and the number of classes), but is linear in relation to the number of training examples \citep{bottou_etal_2018}.  Further, the rate of convergence is not determined by the number of training examples.  For large datasets, convergence may occur in a single pass of the data, or even without using all of the training data (\citealt[pp. 286--288]{goodfellow_etal_2016}; \citealt{bottou_etal_2018}).

\paragraph{Ridge regression.}

In practice, the ridge regression classifier is significantly faster than logistic regression on smaller datasets because it can make use of so-called generalized cross-validation to determine appropriate regularization.  The implementation used here employs eigen decomposition where there are more features than training examples, or singular value decomposition otherwise, with effective complexity of $O(n^{2} \cdot f)$ and $O(n \cdot f^{2})$ respectively \citep[see][]{dongarra_etal_2018}, where $n$ is the number of training examples and $f$ is the number of features.

This makes the ridge regression classifier less scalable for large datasets.  This also requires the complete transform, and does not work incrementally.  In practice, these limitations do not affect any of the datasets in the UCR archive.  For larger datasets, where the importance of regularization decreases, and it is appropriate to perform the transform incrementally, the benefit of using the ridge regression classifier wanes, and training with stochastic gradient descent makes more sense.

\section{Experiments} \label{section-experiments}

We evaluate \textsc{Rocket} on the UCR archive (section \ref{subsection-ucr}), demonstrating that \textsc{Rocket} is competitive with current state-of-the-art methods, obtaining the best mean rank over the 85 `bake off' datasets.

We evaluate scalability in terms of both training set size and time series length (section \ref{subsection-scalability}), demonstrating that \textsc{Rocket} is orders of magnitude faster than current methods.  We also evaluate the effect of different kernel parameters (section \ref{subsection-sensitivity-analysis}), showing that several alternative configurations of \textsc{Rocket} perform similarly well, which is a good indication of the power of the idea, rather than of its fine-tuning.  Unless otherwise stated, all experiments use 10,000 kernels.

The experiments on the datasets in the UCR archive are performed using \textsc{Rocket} in conjunction with a ridge regression classifier, and the experiment in relation to training set size is performed using \textsc{Rocket} integrated with logistic regression.  The experiments on the UCR archive were conducted on a cluster (but using a single CPU core per experiment, not parallelised for speed).  The experiments in relation to scalability (both time series length and training set size) were performed locally using an Intel Core i5-5200U dual-core processor.

\subsection{UCR} \label{subsection-ucr}

\subsubsection{`Bake Off' Datasets}

We evaluate \textsc{Rocket} on the 85 `bake off' datasets from the UCR archive (on the original training/test split for each dataset).  The results presented for \textsc{Rocket} are mean results over 10 runs (using a different set of random kernels for each run).

We compare \textsc{Rocket} to existing state-of-the-art methods for time series classification, namely, BOSS, Shapelet Transform, Proximity Forest, ResNet, and HIVE-COTE.  We also compare \textsc{Rocket} with two more recent methods (with papers on arXiv), InceptionTime and TS-CHIEF, that have been demonstrated to be competitive with HIVE-COTE, while being more scalable.  The results for BOSS, Shapelet Transform, and HIVE-COTE are taken from \citet{bagnall_etal_2019}.

For comparability with other published results, we compare \textsc{Rocket} to the other methods on all 85 `bake off' datasets.  However, as noted above, \textsc{Rocket} was developed using a subset of 40 randomly-selected datasets, to make sure we didn't overfit the UCR archive.  Separate rankings for the 40 `development' datasets, as well as the remaining 45 `holdout' datasets, are provided in Appendix \ref{section-appendix-rank-development-holdout}.

Figure~\ref{figure-rank-ucr85} on page~\pageref{figure-rank-ucr85} gives the mean rank for each method included in the comparison.  Classifiers for which the difference in pairwise classification accuracy is not statistically significant, as determined by a Wilcoxon signed-rank test with Holm correction (as a post hoc test to the Friedman test), are connected with a black line \citep[see][]{demsar_2006,garcia_and_herrera_2008,benavoli_etal_2016}.  The relative accuracy of \textsc{Rocket} and each of the other methods included in the comparison is shown in Figure~\ref{figure-scatterplots-ucr85}, Appendix \ref{section-appendix-relative-accuracy}.

Figure~\ref{figure-rank-ucr85} on page~\pageref{figure-rank-ucr85} shows that \textsc{Rocket} is competitive with (in fact, ranks slightly ahead of) HIVE-COTE, TS-CHIEF, and InceptionTime, although the difference in accuracy between \textsc{Rocket}, HIVE-COTE, InceptionTime, and TS-CHIEF is not statistically significant.  TS-CHIEF ranks ahead of \textsc{Rocket} on the 45 `holdout' datasets (Figure~\ref{figure-rank-ucr45}, Appendix \ref{section-appendix-rank-development-holdout}), but the difference is not significant.

\subsubsection{Additional 2018 Datasets}

\begin{figure}
\centering
\vspace*{-1mm}
\includegraphics[width=0.35\linewidth]{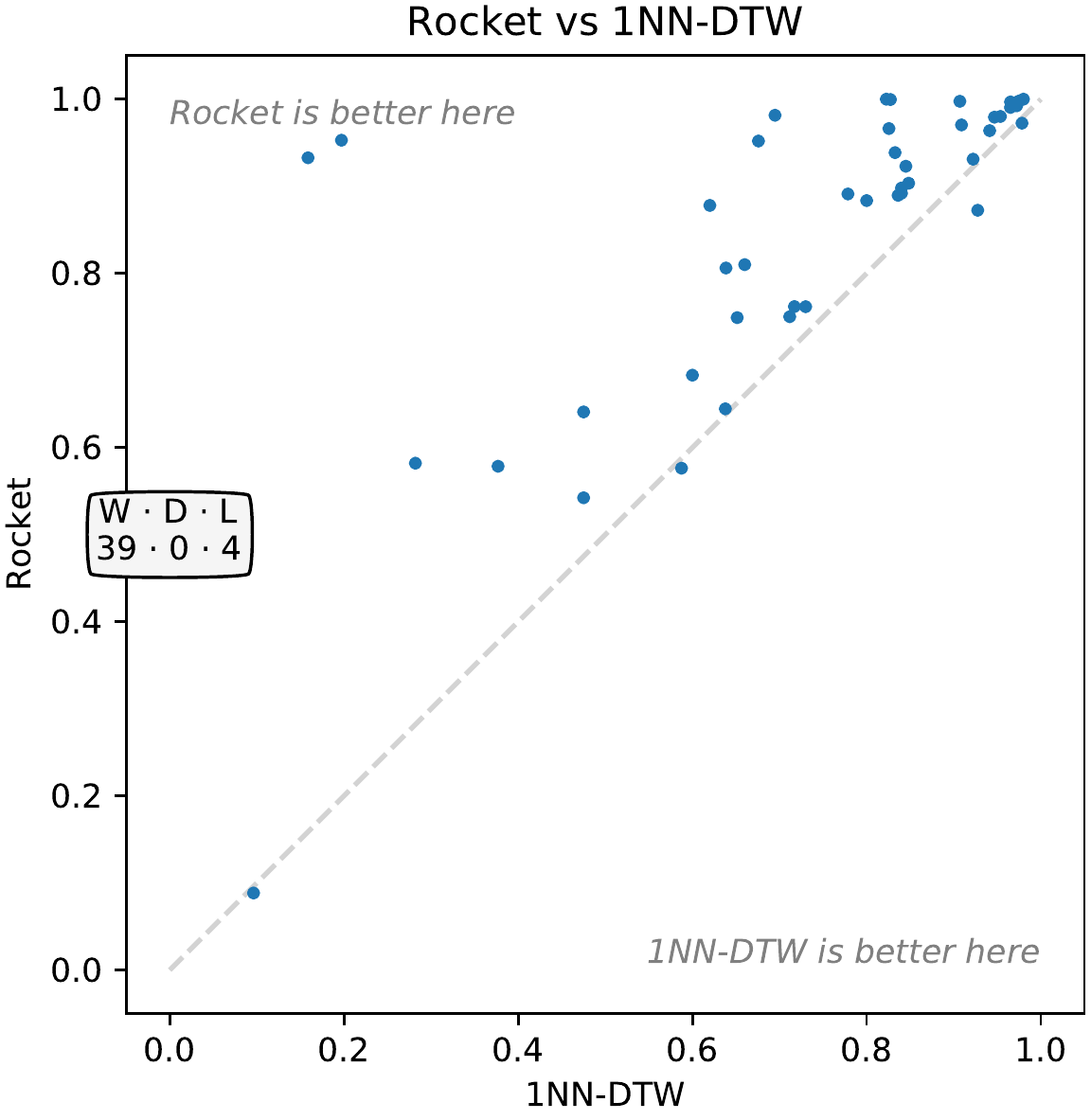}
\vspace*{-1mm}
\caption{Relative accuracy of \textsc{Rocket} versus 1NN-DTW on the 43 additional 2018 datasets.}
\label{figure-scatter-ucr43}
\end{figure}

We have also evaluated \textsc{Rocket} on the 43 additional datasets in the UCR archive as of 2018, in order to: (1) show that our method is able to handle datasets with varying lengths; and (2) provide reference results for future research papers.

There are no published results for state-of-the-art methods on these datasets.  Adapting these methods to work on variable-length time series is nontrivial, as the most appropriate method for handling variable lengths (1) depends on whether the variable lengths represent subsampling or variable sampling frequencies, and (2) is classifier dependent \citep{tan_etal_2019}.  Accordingly, we restrict our comparison to the available results for 1NN-DTW \citep{dau_etal_2019}, where variable length time series have been padded with `low amplitude random [noise]' to the same length as the longest time series \citep[p. 16]{dau_etal_2018}.  Figure~\ref{figure-scatter-ucr43} shows the relative accuracy of \textsc{Rocket} and 1NN-DTW on the 43 additional datasets.  \textsc{Rocket} is more accurate on all but four datasets, and substantially more so on most.

Following \citet{dau_etal_2018}, we have normalized each time series and interpolated missing values.  Variable-length time series have been rescaled or used `as is' (with their original lengths) as determined by 10-fold cross-validation.

\subsection{Scalability} \label{subsection-scalability}

\subsubsection{Training Set Size}

\begin{figure}
\centering
\vspace*{-1mm}
\includegraphics[width=0.8\linewidth]{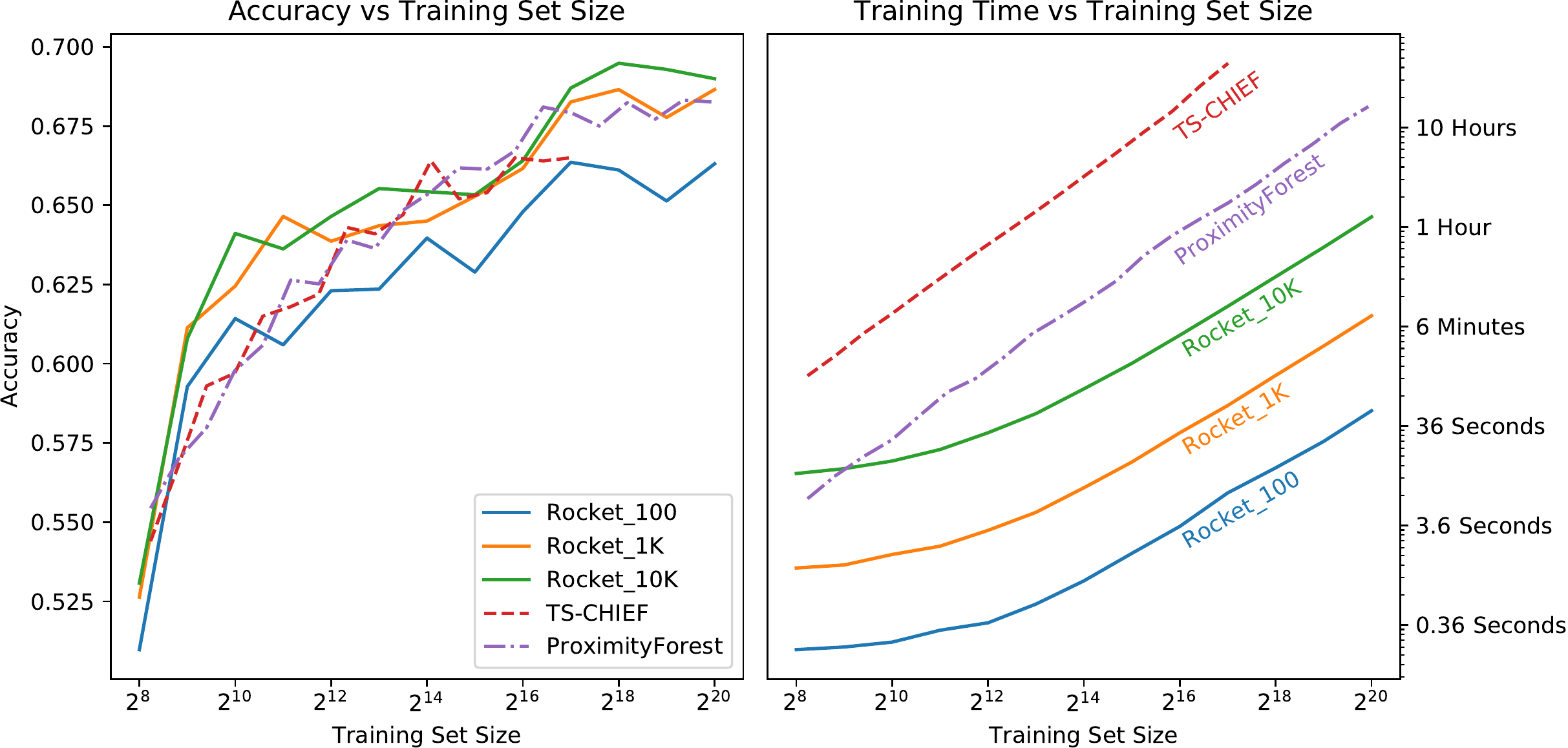}
\vspace*{-1mm}
\caption{Accuracy (left) and training time (right) versus training set size.}
\label{figure-scalability-satellite}
\end{figure}

Following \citet{lucas_etal_2019}, \citet{shifaz_etal_2019}, and \citet{ismailfawaz_etal_2019_c}, we evaluate scalability in terms of training set size on increasingly larger subsets (up to approximately 1 million time series) of the Satellite Image Time Series dataset \citep[see][]{petitjean_etal_2012}.  The time series in this dataset represent a vegetation index, calculated from spectral data acquired by the Formosat-2 satellite, and the classes represent different land cover types.  The aim in classifying these time series is to map different vegetation profiles to different types of crops and forested areas.  Each time series has a length of 46.

For this purpose, we integrate \textsc{Rocket} with logistic regression.  The transform is performed in tranches, which are further divided into minibatches for training.  Each time series is normalised to have a zero mean and unit standard deviation.

We train the model for at least one epoch for each subset size.  To prevent overfitting, we stop training (after the first epoch) if validation loss has failed to improve after 20 updates.  In practice, while training may continue for 40 or 50 epochs for smaller subset sizes, training converges within a single pass for anything more than approximately 16,000 training examples.  Validation loss is computed on a separate validation set (the same set of 2,048 examples for all subset sizes).

Optimization is performed using Adam \citep{kingma_and_ba_2015}.  We perform a minimal search on the initial learning rate to ensure that training loss does not diverge.  The learning rate is halved if training loss fails to improve after 100 updates (only relevant for larger subset sizes).

We have run \textsc{Rocket} in three guises: with 100, 1,000, and 10,000 kernels (the default).  We compare \textsc{Rocket} against Proximity Forest and TS-CHIEF, which have already been demonstrated to be fundamentally more scalable than HIVE-COTE \citep{lucas_etal_2019,shifaz_etal_2019}.  (Results for larger quantities of data are not yet available for InceptionTime.)

Figure~\ref{figure-scalability-satellite} shows classification accuracy and training time versus training set size for \textsc{Rocket}, Proximity Forest, and TS-CHIEF.  As expected, \textsc{Rocket} scales linearly with respect to both the number of training examples, and the number of kernels (see section \ref{subsection-complexity-analysis}).  With 1,000 or 10,000 kernels, \textsc{Rocket} achieves similar classification accuracy to Proximity Forest and TS-CHIEF.  With 100 kernels, \textsc{Rocket} achieves lower classification accuracy, but takes less than a minute to learn from more than 1 million time series.  Even with 10,000 kernels, \textsc{Rocket} is an order of magnitude faster than Proximity Forest.  (Training time for smaller subset sizes for \textsc{Rocket} is dominated by the cost of the transform for the validation set, which is why training time is `flat' for smaller subset sizes.)

\subsubsection{Time Series Length}

\begin{figure}
\centering
\vspace*{-1mm}
\includegraphics[width=0.5\linewidth]{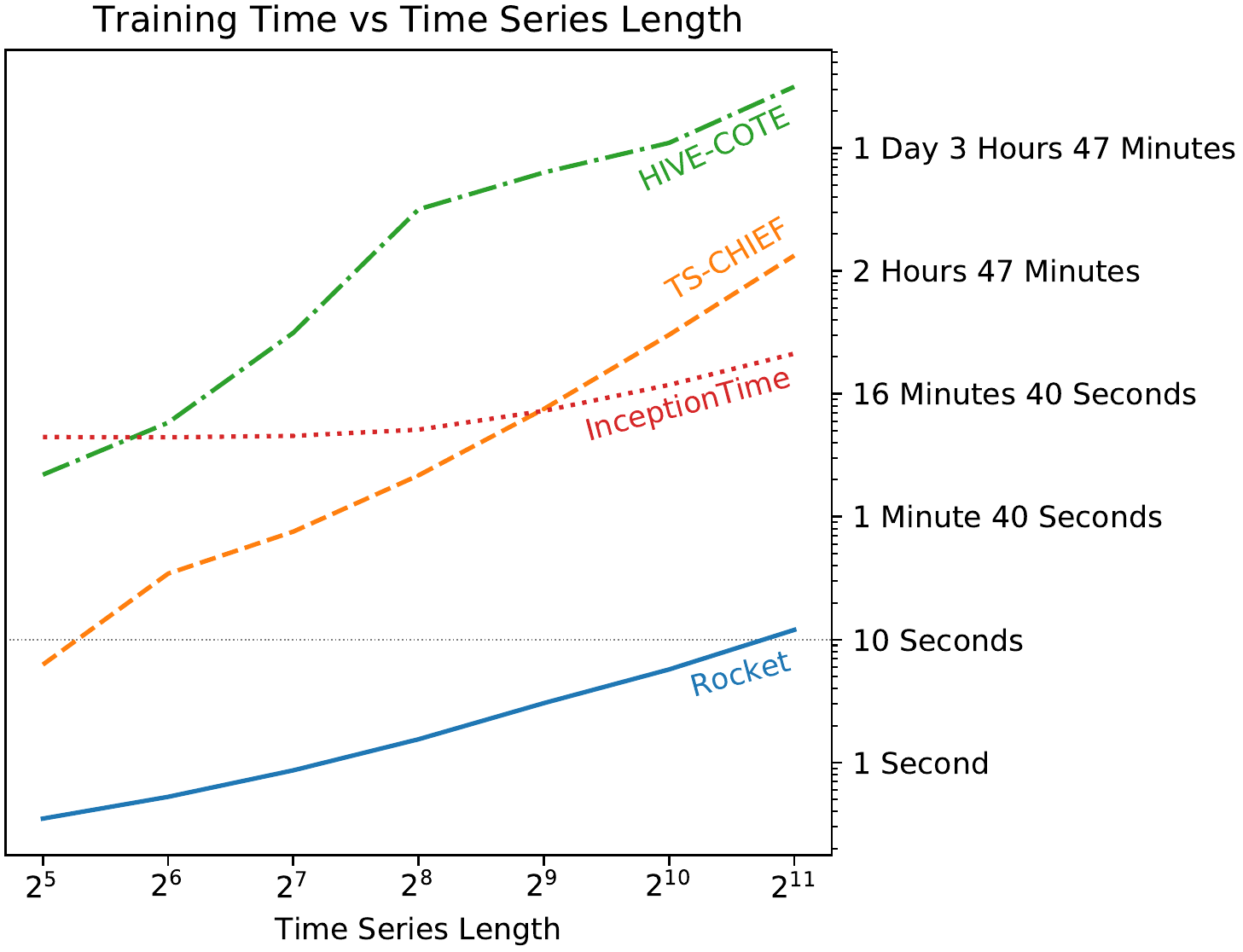}
\vspace*{-1mm}
\caption{Training time versus time series length.}
\label{figure-scalability-length}
\end{figure}

Following \citet{shifaz_etal_2019} and \citet{ismailfawaz_etal_2019_c}, we evaluate scalability in terms of time series length using the \textit{InlineSkate} dataset from the UCR archive.  We use \textsc{Rocket} in the same configuration as for the other datasets in the UCR archive (that is, using the ridge regression classifier and 10,000 kernels).  Results for HIVE-COTE and TS-CHIEF are taken from \citet{shifaz_etal_2019}.

Figure~\ref{figure-scalability-length} shows training time versus time series length for \textsc{Rocket}, TS-CHIEF, HIVE-COTE, and InceptionTime.  The difference in training time between \textsc{Rocket} and TS-CHIEF is substantial.  \textsc{Rocket} takes approximately as long to train on time series of length 2,048 as TS-CHIEF takes for time series of length 32, and is approximately three orders of magnitude faster for the longest time series length.

\textsc{Rocket} is considerably faster than InceptionTime as well.  However, fundamental scalability is likely to be similar, given that both InceptionTime and \textsc{Rocket} are based on convolutional architectures.

\subsection{Sensitivity Analysis} \label{subsection-sensitivity-analysis}

We explore the effect of different kernel parameters on classification accuracy.  We compare the accuracy of the default configuration (i.e., using the parameters specified in section \ref{section-method}) against different choices for the number of kernels, length, weights and bias, dilation, padding, and output features.  In each case, only the given parameter (e.g., length) is varied, keeping all other parameters fixed at their default values.  The comparison is made on the `development' datasets.  The results are mean results over 10 runs (using a different set of random kernels per run).

In most cases, alternative configurations represent a relatively subtle change from the default configuration.  Unsurprisingly, therefore, in many cases one or more alternative choices for the relevant parameter produces similar accuracy to the baseline configuration.  In other words, \textsc{Rocket} is relatively robust to different choices for many parameters.  However, it is clear that dilation and \textit{ppv}, in particular, are two key aspects of the performance of the method.

\subsubsection{Number of Kernels} \label{subsubsection-number-of-kernels}

\begin{figure}
  \centering
  \vspace*{-1mm}
  \subfloat{
    \includegraphics[width=0.65\linewidth]{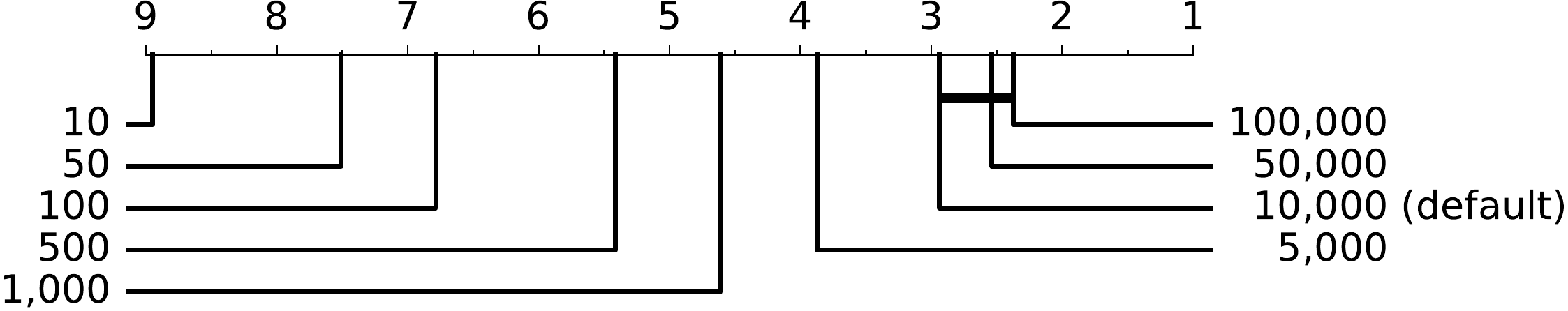}
  }
  \subfloat{
    \includegraphics[width=0.3\linewidth]{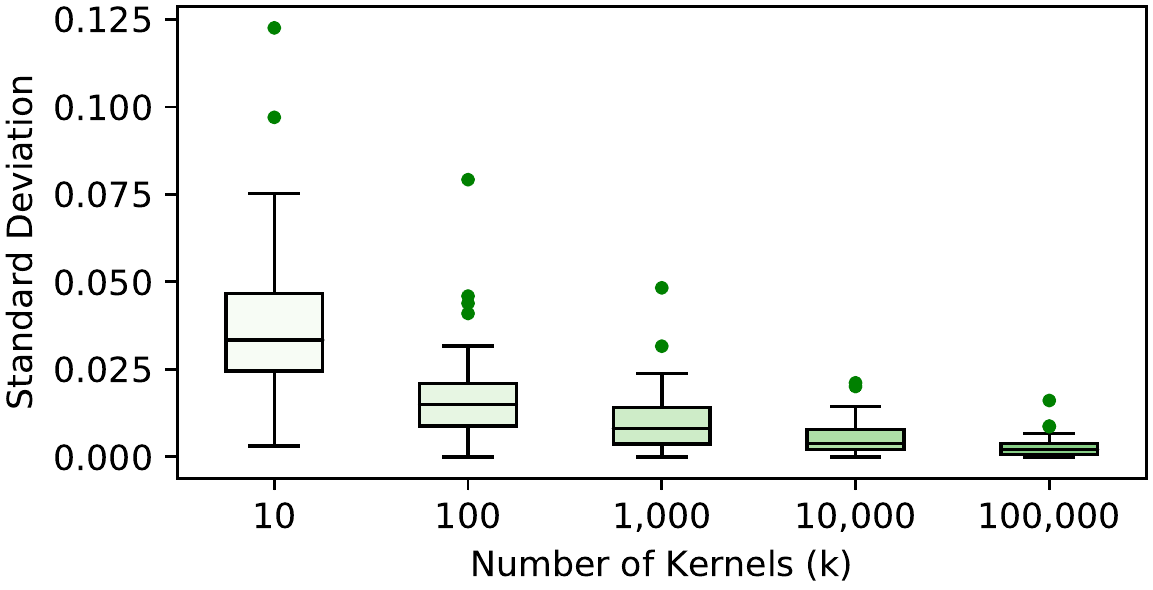}
  }
  \vspace*{-1mm}
  \caption{Mean ranks (left), and variance in accuracy (right), versus $k$.}
  \label{figure-rank-num-kernels}
\end{figure}

We evaluate increasing numbers of kernels between 10 and 100,000.  Figure~\ref{figure-rank-num-kernels} shows the effect of the number of kernels, $k$, on accuracy.  Clearly, increasing the number of kernels improves accuracy.  However, the actual difference in accuracy between, for example, $k=5{,}000$ and $k=10{,}000$, is relatively small, even if statistically significant.  Indeed, $k=5{,}000$ produces higher accuracy on some datasets (Figure~\ref{fig-scatter-num-kernels}, Appendix \ref{section-appendix-additional-plots}).  Nevertheless, $k=10{,}000$ is noticeably ahead in terms of win/draw/loss (29/3/8).  The differences between $k=10{,}000$, $k=50{,}000$, and $k=100{,}000$ are not statistically significant.

Even though \textsc{Rocket} is nondeterministic, the variability in accuracy is reasonably low for large numbers of kernels.  Unsurprisingly, standard deviation diminishes as $k$ increases.  The median standard deviation across the 40 `development' datasets is 0.0038 for $k=10{,}000$, and 0.0021 for $k=100{,}000$.

\subsubsection{Kernel Length}

\begin{figure}
\centering
\vspace*{-1mm}
\includegraphics[width=0.9\linewidth]{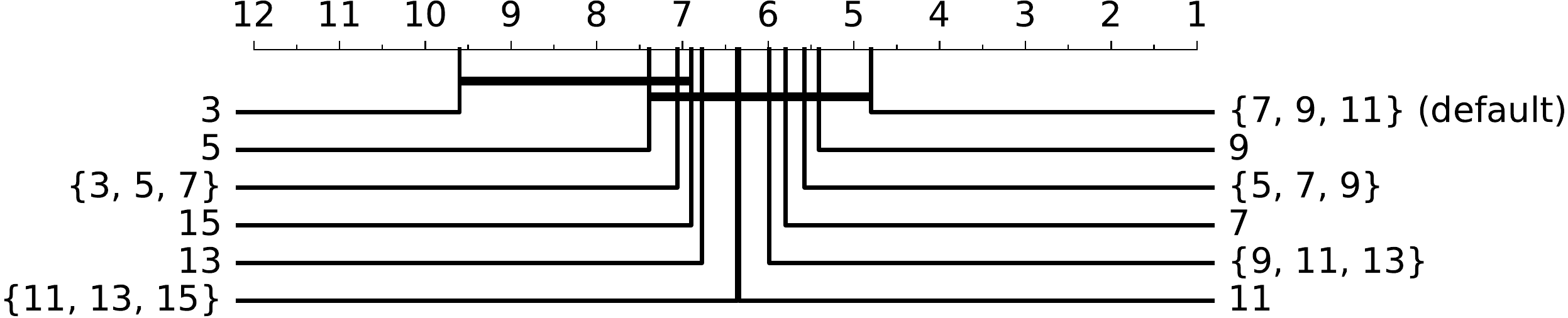}
\vspace*{-1mm}
\caption{Mean ranks for different choices in terms of kernel length.}
\label{figure-rank-ablation-length}
\end{figure}

We vary kernel length, comparing the baseline (selecting length randomly from $\{7, 9, 11\}$) to:

\begin{itemize}
  \item fixed lengths of 3, 5, 7, 9, 11, 13, and 15; and
  \item selecting length randomly from $\{3, 5, 7\}$, $\{5, 7, 9\}$, $\{9, 11, 13\}$, and $\{11, 13, 15\}$.
\end{itemize}

Figure~\ref{figure-rank-ablation-length} shows the effect of these choices on accuracy.  Fixed lengths of 7, 9, and 11, as well as selecting length randomly from $\{5, 7, 9\}$ and $\{9, 11, 13\}$ result in similar accuracy to the default configuration, and the differences are not statistically significant (see also Figure~\ref{figure-scatter-ablation-length}, Appendix \ref{section-appendix-additional-plots}).  Shorter kernels are undesirable, being more strongly correlated with each other for a large number of kernels.

\subsubsection{Weights (Including Centering) and Bias}

\paragraph{Weights.}

\begin{figure}
\centering
\vspace*{-1mm}
\includegraphics[width=0.9\linewidth]{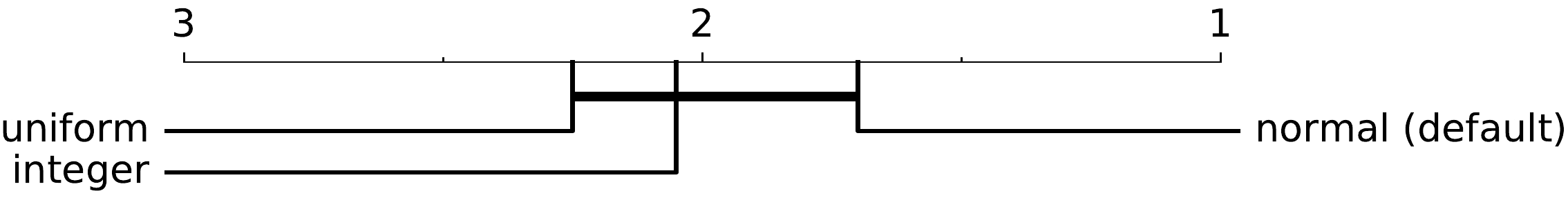}
\vspace*{-1mm}
\caption{Mean ranks for different choices in terms of the sampling distribution for the weights.}
\label{figure-rank-ablation-weights}
\end{figure}

We vary the distribution from which the weights are sampled, comparing the baseline (sampling from a normal distribution) to:

\begin{itemize}
  \item sampling from a uniform distribution, $\forall w \in \boldsymbol{W}$, $w~\sim~\mathcal{U}(-1, 1)$; and
  \item sampling integer weights uniformly from $\{-1, 0, 1\}$.
\end{itemize}

Figure~\ref{figure-rank-ablation-weights} shows the effect of these choices on accuracy.  While sampling from a normal distribution produces higher accuracy, the actual difference in accuracy is small and not statistically significant (see also Figure~\ref{figure-scatter-ablation-weights}, Appendix \ref{section-appendix-additional-plots}).  While it may seem surprising that weights sampled from only three integer values are so effective, note that kernels are still mean centered by default and have random bias, and there is still substantial variety in terms of length and dilation.

\paragraph{Centering.}

\begin{figure}
\centering
\vspace*{-1mm}
\includegraphics[width=0.9\linewidth]{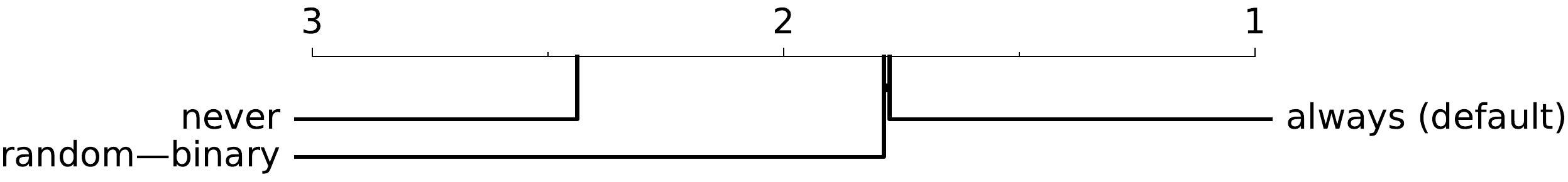}
\vspace*{-1mm}
\caption{Mean ranks for different choices in terms of centering.}
\label{figure-rank-ablation-centering}
\end{figure}

We vary centering, comparing the baseline (always centering) against:

\begin{itemize}
  \item never centering the kernel weights; and
  \item centering or not centering at random with equal probability.
\end{itemize}

Figure~\ref{figure-rank-ablation-centering} shows the effect of these choices on accuracy.  It is clear that centering produces higher accuracy, but the difference between always centering and centering at random is very small and not statistically significant.  Always centering, however, is noticeably more accurate on some datasets (Figure~\ref{figure-scatter-ablation-centering}, Appendix \ref{section-appendix-additional-plots}).

\paragraph{Bias.}

\begin{figure}
\centering
\vspace*{-1mm}
\includegraphics[width=0.9\linewidth]{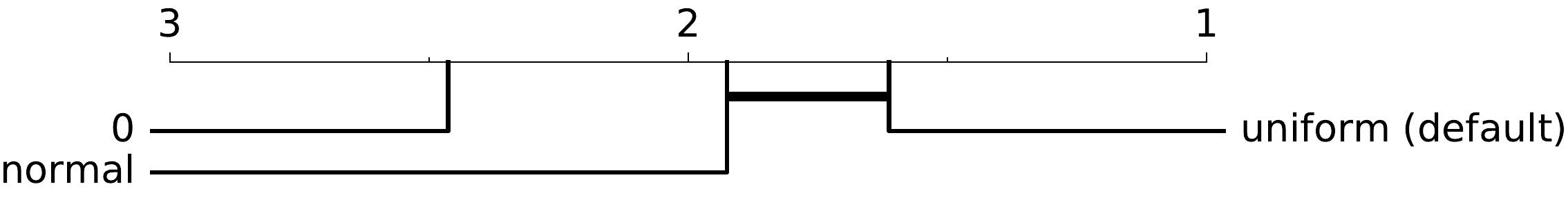}
\vspace*{-1mm}
\caption{Mean ranks for different choices in terms of bias.}
\label{figure-rank-ablation-bias}
\end{figure}

We vary bias, comparing the baseline (using a uniform distribution) against:

\begin{itemize}
  \item using zero bias; and
  \item sampling bias from a normal distribution, $b \sim \mathcal{N}(0, 1)$.
\end{itemize}

Figure~\ref{figure-rank-ablation-bias} shows the effect of these choices on accuracy.  Using a bias term produces higher accuracy, but the difference between sampling bias from a uniform distribution or a normal distribution is relatively small and not statistically significant (see also Figure~\ref{figure-scatter-ablation-bias}, Appendix \ref{section-appendix-additional-plots}.)

\subsubsection{Dilation} \label{subsubsection-dilation}

\begin{figure}
\centering
\vspace*{-1mm}
\includegraphics[width=0.9\linewidth]{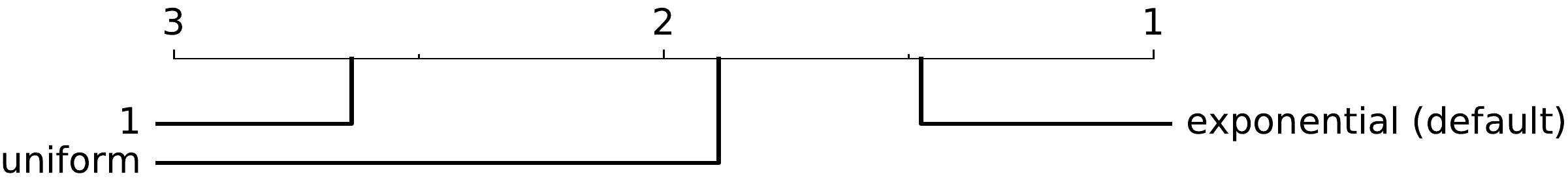}
\vspace*{-1mm}
\caption{Mean ranks for different choices in terms of dilation.}
\label{figure-rank-ablation-dilation}
\end{figure}

We vary dilation, comparing the baseline (sampling dilation on an exponential scale) against:

\begin{itemize}
  \item no dilation (i.e., a fixed dilation of one); and
  \item sampling dilation uniformly, $d = \lfloor x \rfloor$, $x \sim \mathcal{U}(1, \frac{l_{\text{input}} - 1}{l_{\text{kernel}} - 1})$.
\end{itemize}

Figure~\ref{figure-rank-ablation-dilation} shows the effect of these choices in terms of accuracy.  It is clear that dilation is key to performance.  Dilation produces obviously higher accuracy than no dilation.  Exponential dilation produces higher accuracy than uniform dilation on most datasets (significantly higher on some datasets), and the difference is statistically significant (see also Figure~\ref{figure-scatter-ablation-dilation}, Appendix \ref{section-appendix-additional-plots}).

\subsubsection{Padding}

\begin{figure}
\centering
\vspace*{-1mm}
\includegraphics[width=0.9\linewidth]{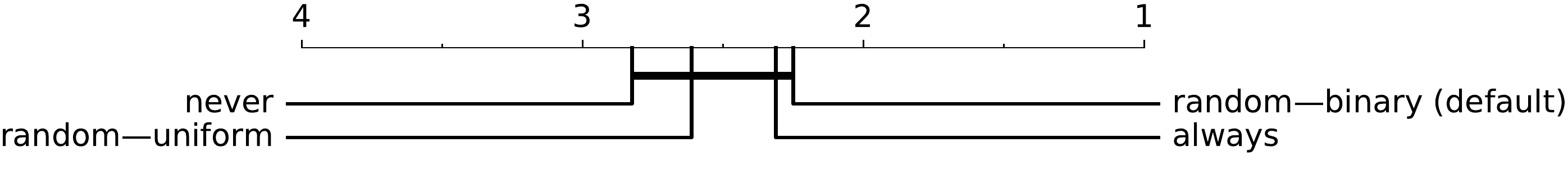}
\vspace*{-1mm}
\caption{Mean ranks for different choices in terms of padding.}
\label{figure-rank-ablation-padding}
\end{figure}

We vary padding, comparing the baseline (applying padding at random) against:

\begin{itemize}
  \item always padding, such that the `middle' element of a given kernel is centered on the first element of the time series, $p = ((l_{\text{kernel}} - 1) \times d) / 2$;
  \item sampling padding uniformly, $p \sim \mathcal{U}(0, ((l_{\text{kernel}} - 1) \times d) / 2)$; and
  \item never padding.
\end{itemize}

Figure~\ref{figure-rank-ablation-padding} shows the effect of these choices on accuracy.  Padding is superior to not padding, but none of the differences are statistically significant.  Different choices produce very similar results (Figure~\ref{figure-scatter-ablation-padding}, Appendix \ref{section-appendix-additional-plots}).

\subsubsection{Features} \label{subsubsection-features}

\begin{figure}
  \centering
  \vspace*{-1mm}
  \subfloat{
    \includegraphics[width=0.65\linewidth]{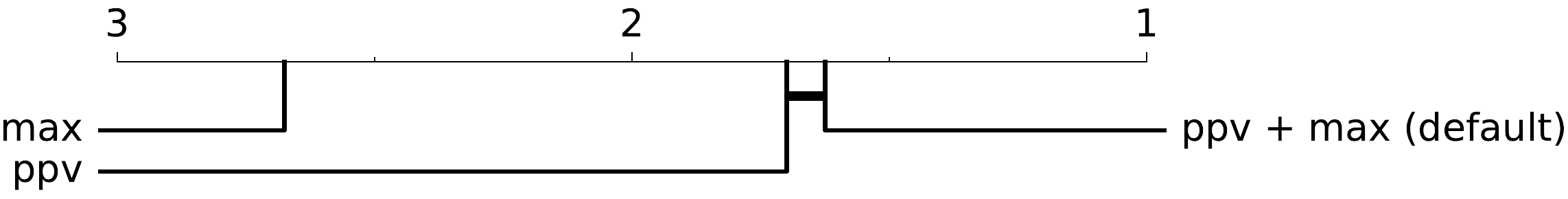}
  }
  \subfloat{
    \includegraphics[width=0.3\linewidth]{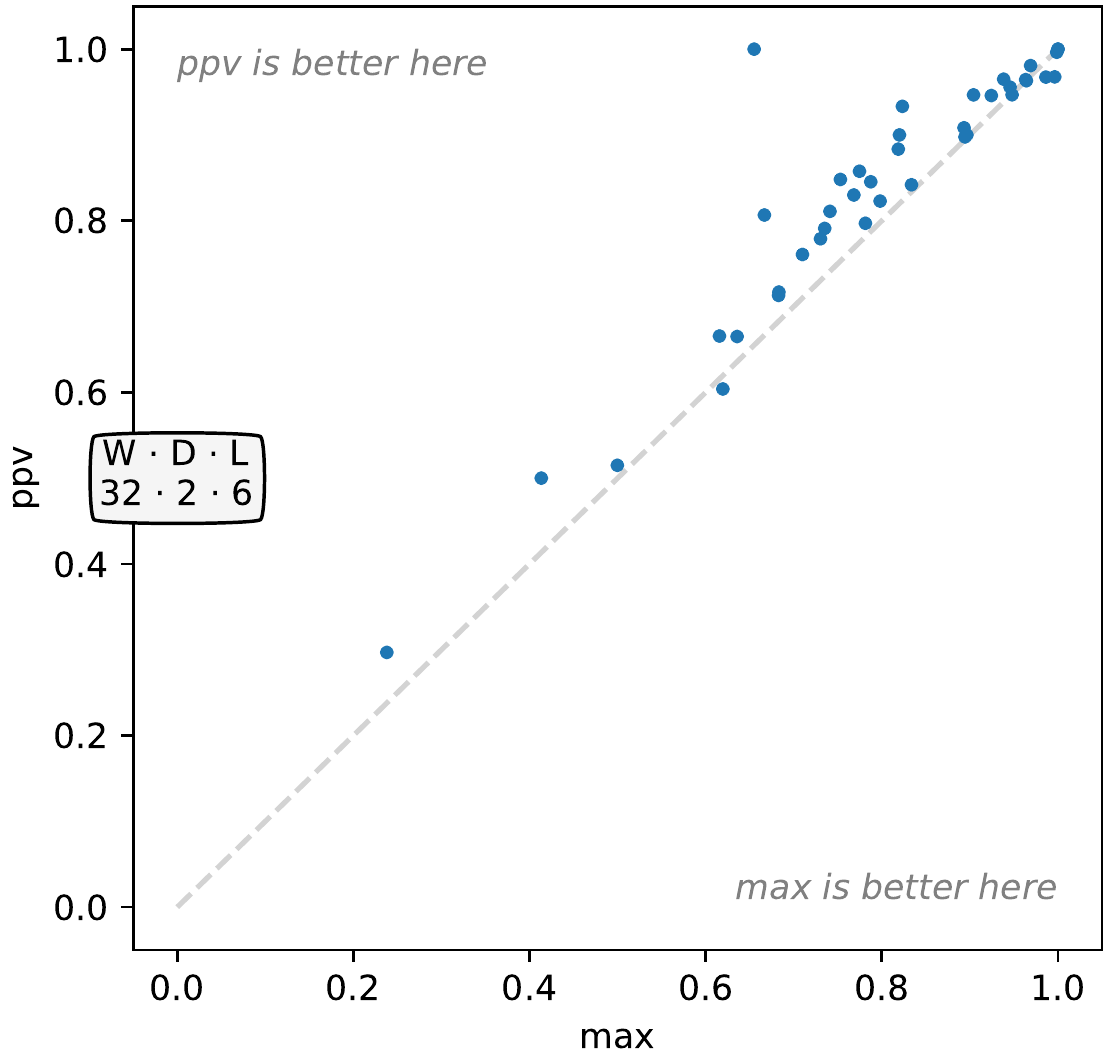}
  }
  \vspace*{-1mm}
  \caption{Mean ranks (left), and relative accuracy (right), \textit{ppv} and max.}
  \label{figure-rank-ablation-features}
\end{figure}

We vary the output features, comparing the baseline, \textit{ppv} and max, against using each in isolation.  Figure~\ref{figure-rank-ablation-features} shows the effect of these choices on accuracy.  It is clear that \textit{ppv} is superior to max: \textit{ppv} produces substantially higher classification accuracy for the majority of the `development' datasets.  In fact, \textit{ppv} has the single biggest effect on accuracy of all the parameters.  The combination of \textit{ppv} and max is better again, although the difference between \textit{ppv} and \textit{ppv} plus max is small and not statistically significant (see also Figure~\ref{figure-scatter-ablation-features}, Appendix \ref{section-appendix-additional-plots}).

\section{Conclusion}

Convolutional kernels are a single, powerful instrument which can capture many of the features used by existing methods for time series classification.  We show that, rather than learning kernel weights, a large number of random kernels---while in isolation only approximating relevant patterns---in combination are extremely effective for capturing discriminative patterns in time series.

Further, random kernels have very low computational requirements, making learning and classification extremely fast.  Our proposed method utilising random convolutional kernels for the purposes of transforming and classifying time series, \textsc{Rocket}, achieves state-of-the-art accuracy with a fraction of the computational expense of existing methods.  \textsc{Rocket} also scales to millions of time series.

\textsc{Rocket} makes key use of the proportion of positive values (or \textit{ppv}) to summarise the output of feature maps, allowing a classifier to weight the prevalence of a pattern in a given time series.  To our knowledge, \textit{ppv} has not been used in this way before.  We find that this is substantially more effective than a simple maximum as applied in a conventional max pooling operation.  It is credible that \textit{ppv} would also be effective for other data types such as images.

In future work, we propose to explore feature selection for \textsc{Rocket}, the application of \textsc{Rocket} to multivariate timeseries, the application of \textsc{Rocket} beyond time series data, and the use of aspects of \textsc{Rocket} with learned kernels.

\begin{acknowledgements}
This material is based upon work supported by an Australian Government Research Training Program Scholarship; the Air Force Office of Scientific Research, Asian Office of Aerospace Research and Development (AOARD) under award number FA2386--18--1--4030; and the Australian Research Council under awards DE170100037 and DP190100017.  The authors would like to thank Professor Eamonn Keogh and all the people who have contributed to the UCR time series classification archive.  Figures showing the ranking of different classifiers and variants of \textsc{Rocket} were produced using code from \citet{ismailfawaz_etal_2019_a}.
\end{acknowledgements}

\bibliographystyle{spbasic}
\bibliography{references}

\clearpage

\appendix

\section*{Appendices}

\section{Relative Accuracy} \label{section-appendix-relative-accuracy}

\begin{figure}[h]
\centering
\vspace*{-1mm}
\includegraphics[width=0.65\linewidth]{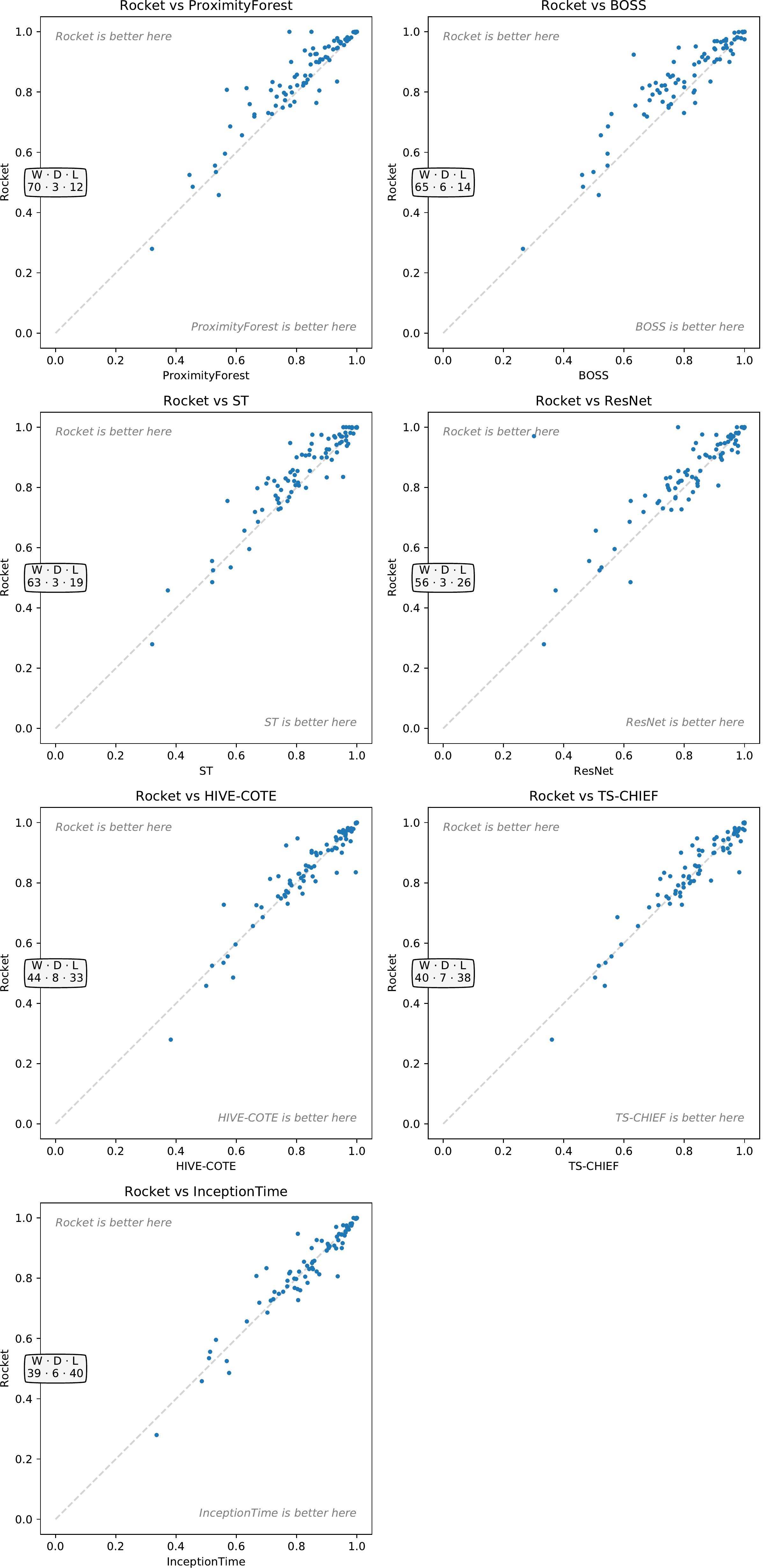}
\vspace*{-1mm}
\caption{Relative accuracy of \textsc{Rocket} vs state-of-the-art classifiers on the `bake off' datasets.}
\label{figure-scatterplots-ucr85}
\end{figure}

\clearpage

\section{`Development' and `Holdout' Datasets} \label{section-appendix-rank-development-holdout}

\begin{figure}[h]
\centering
\vspace*{-1mm}
\includegraphics[width=0.9\linewidth]{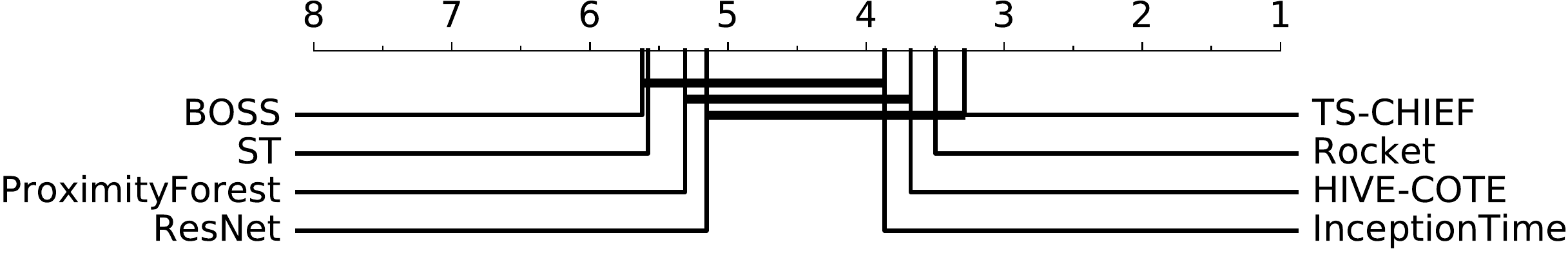}
\vspace*{-1mm}
\caption{Mean rank of \textsc{Rocket} versus state-of-the-art classifiers on the `holdout' datasets.}
\label{figure-rank-ucr45}
\end{figure}

\begin{figure}[h]
\centering
\vspace*{-1mm}
\includegraphics[width=0.9\linewidth]{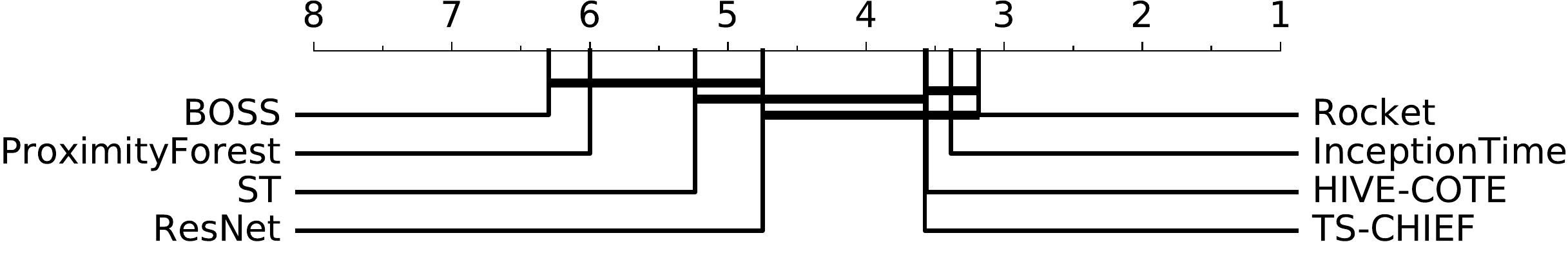}
\vspace*{-1mm}
\caption{Mean rank of \textsc{Rocket} vs state-of-the-art classifiers on the `development' datasets.}
\label{figure-rank-ucr40}
\end{figure}

\section{Additional Plots} \label{section-appendix-additional-plots}

\begin{figure}[h]
\centering
\vspace*{-1mm}
\includegraphics[width=0.35\linewidth]{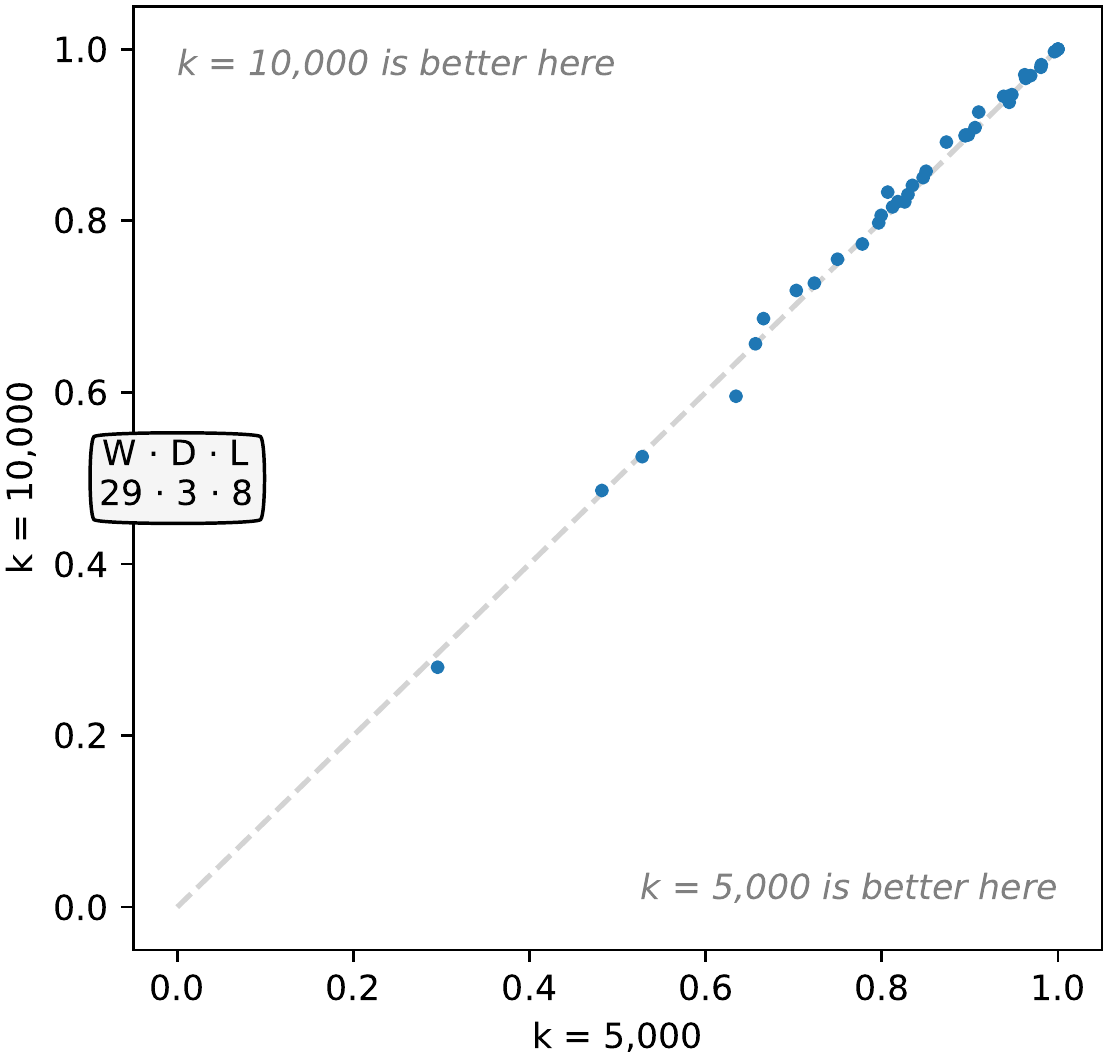}
\vspace*{-1mm}
\caption{Relative accuracy of $k=10{,}000$ versus $k=5{,}000$ on the `development' datasets.}
\label{fig-scatter-num-kernels}
\end{figure}

\begin{figure}[h]
\centering
\vspace*{-1mm}
\includegraphics[width=0.35\linewidth]{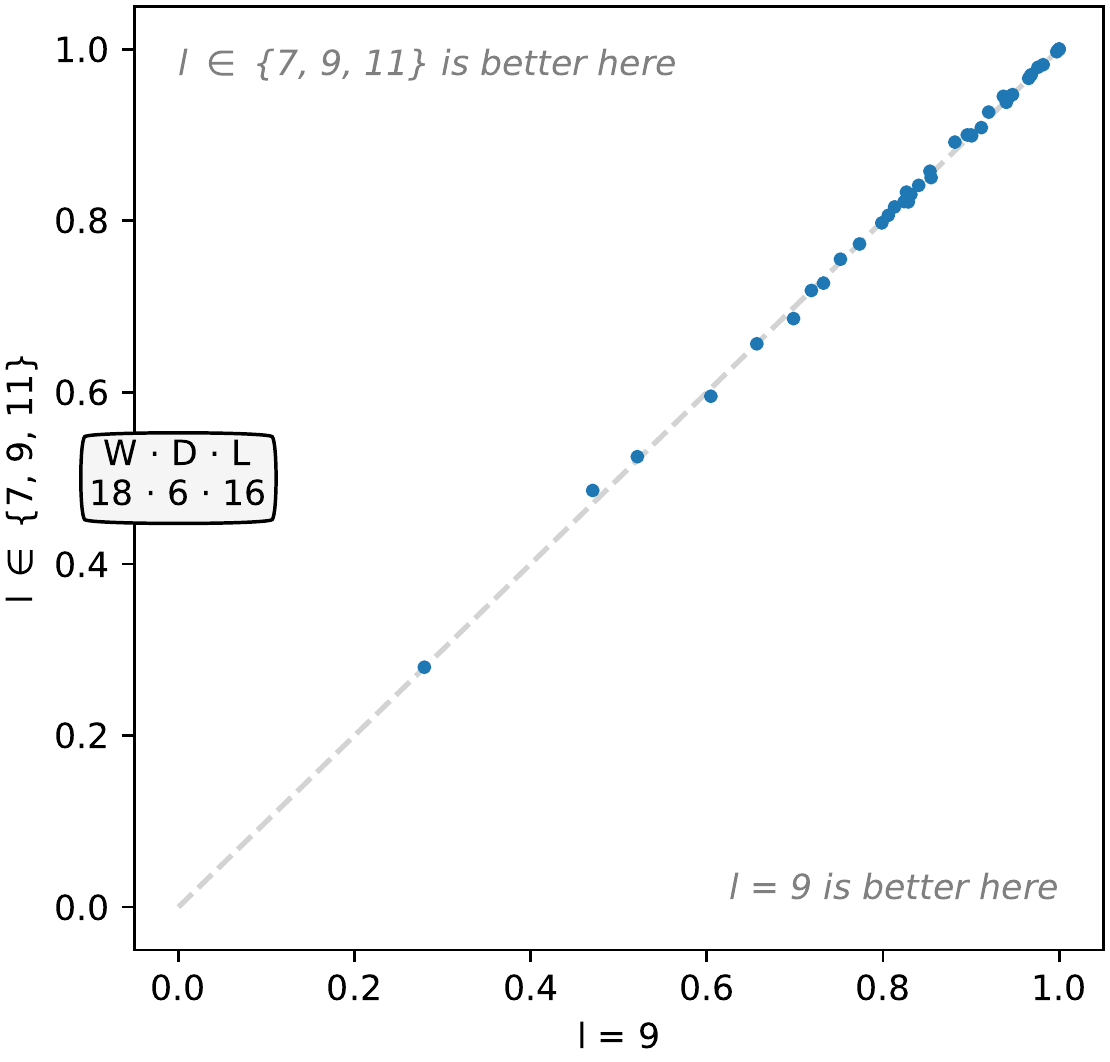}
\vspace*{-1mm}
\caption{Relative accuracy of $l \in \{7, 9, 11\}$ versus $l = 9$ on the `development' datasets.}
\label{figure-scatter-ablation-length}
\end{figure}

\begin{figure}[h]
\centering
\vspace*{-1mm}
\includegraphics[width=0.35\linewidth]{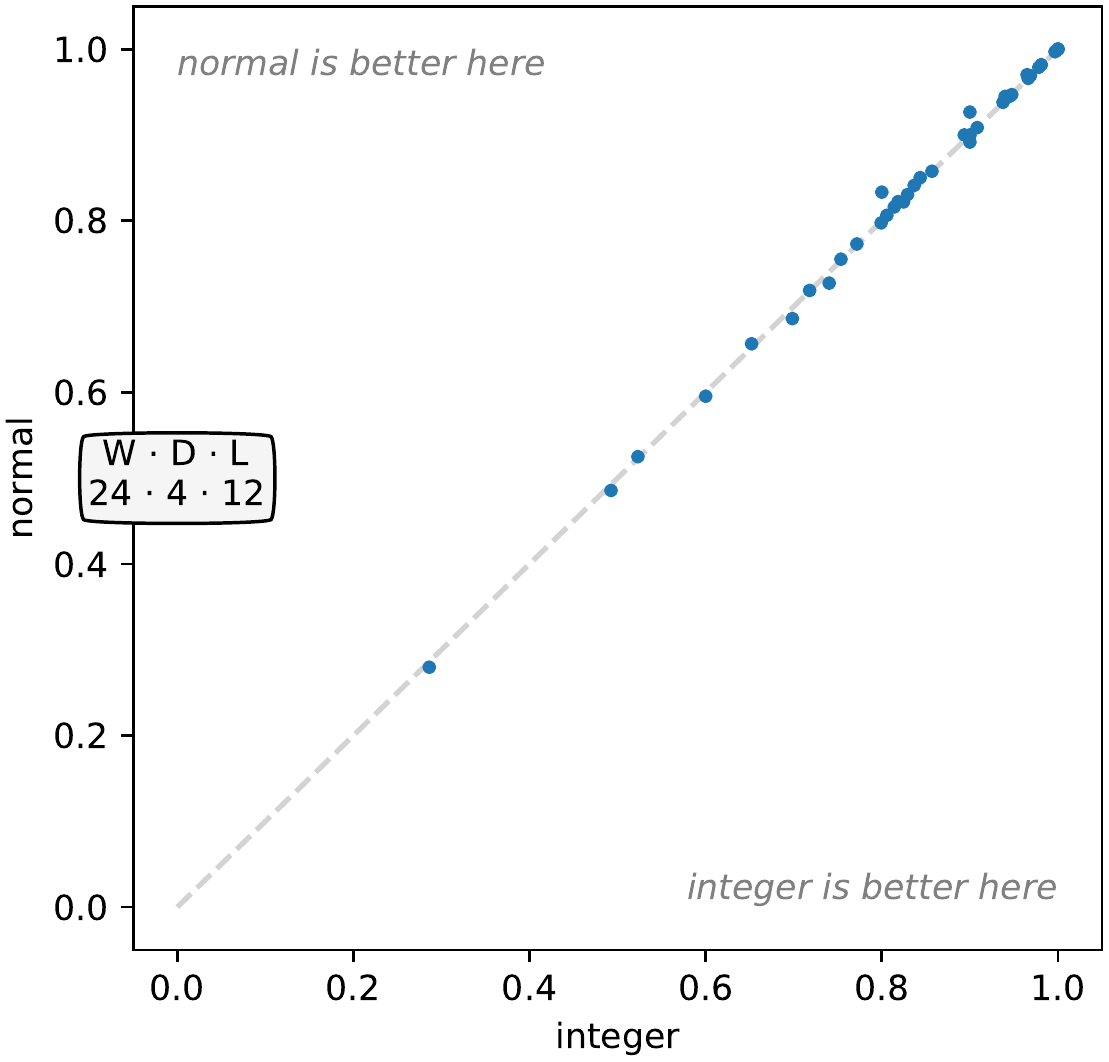}
\vspace*{-1mm}
\caption{Relative accuracy, normally-distributed vs integer weights, `development' datasets.}
\label{figure-scatter-ablation-weights}
\end{figure}

\begin{figure}[h]
\centering
\vspace*{-1mm}
\includegraphics[width=0.35\linewidth]{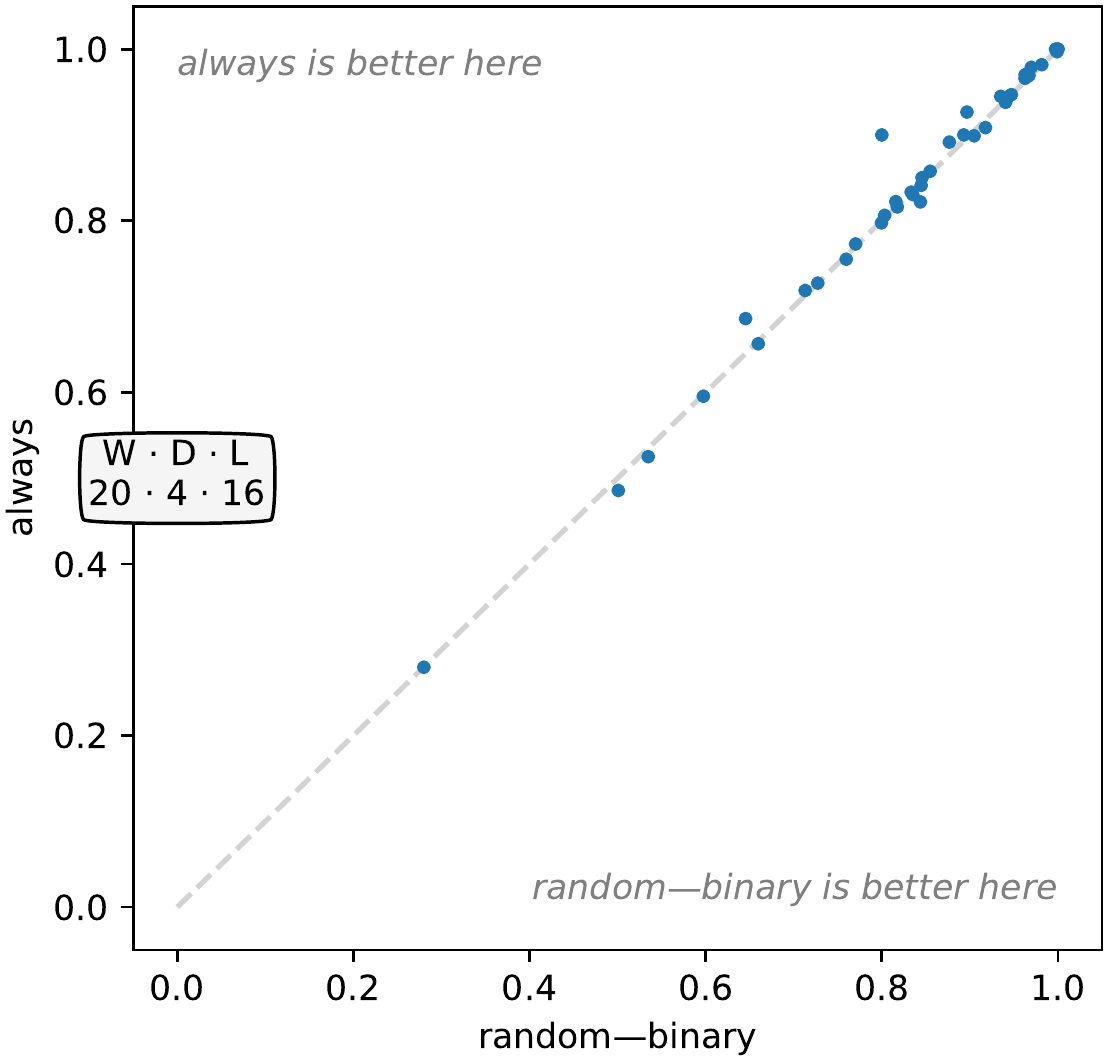}
\vspace*{-1mm}
\caption{Relative accuracy of always vs random centering on the `development' datasets.}
\label{figure-scatter-ablation-centering}
\end{figure}

\begin{figure}[h]
\centering
\vspace*{-1mm}
\includegraphics[width=0.35\linewidth]{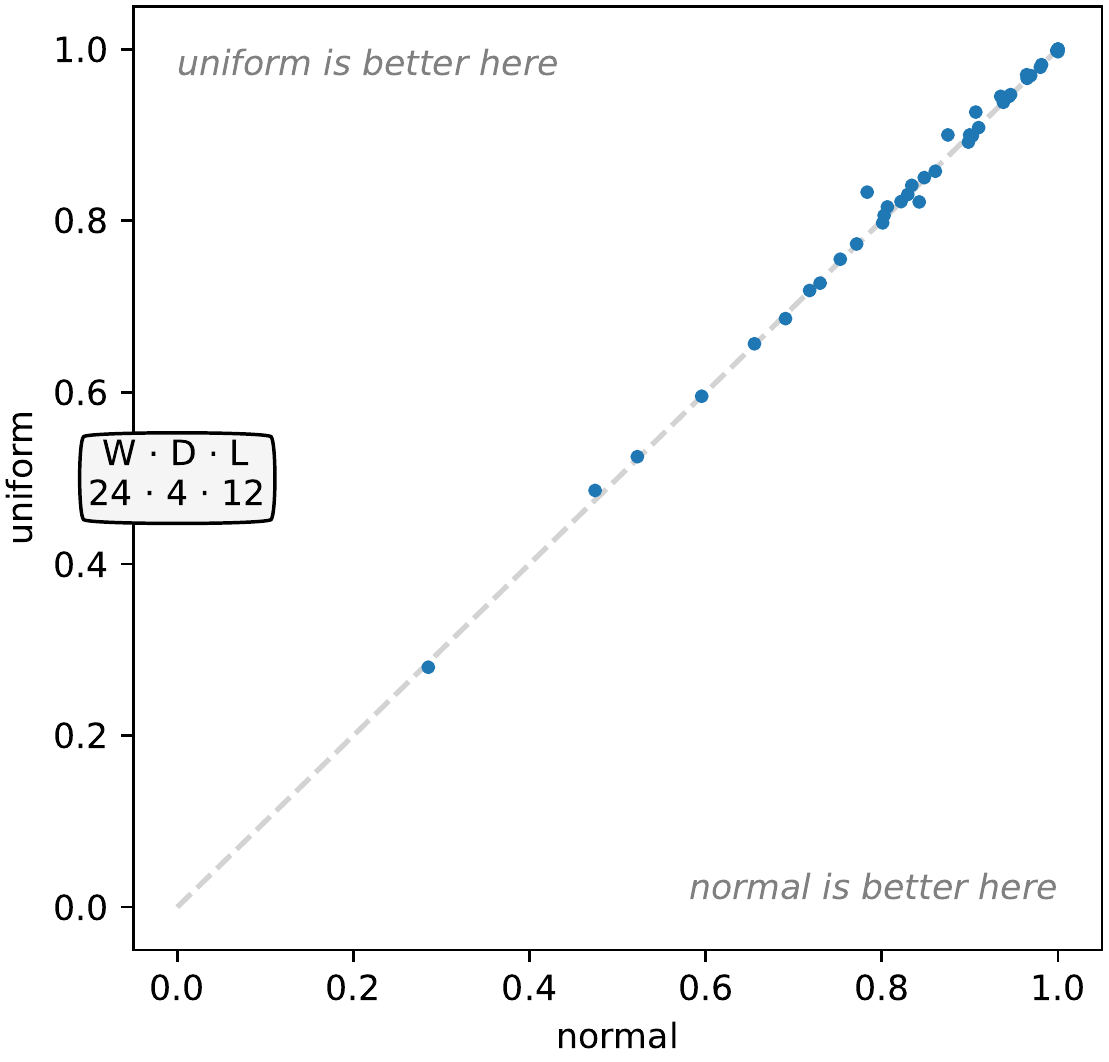}
\vspace*{-1mm}
\caption{Relative accuracy, uniformly versus normally-distributed bias, `development' datasets.}
\label{figure-scatter-ablation-bias}
\end{figure}

\begin{figure}[h]
\centering
\vspace*{-1mm}
\includegraphics[width=0.35\linewidth]{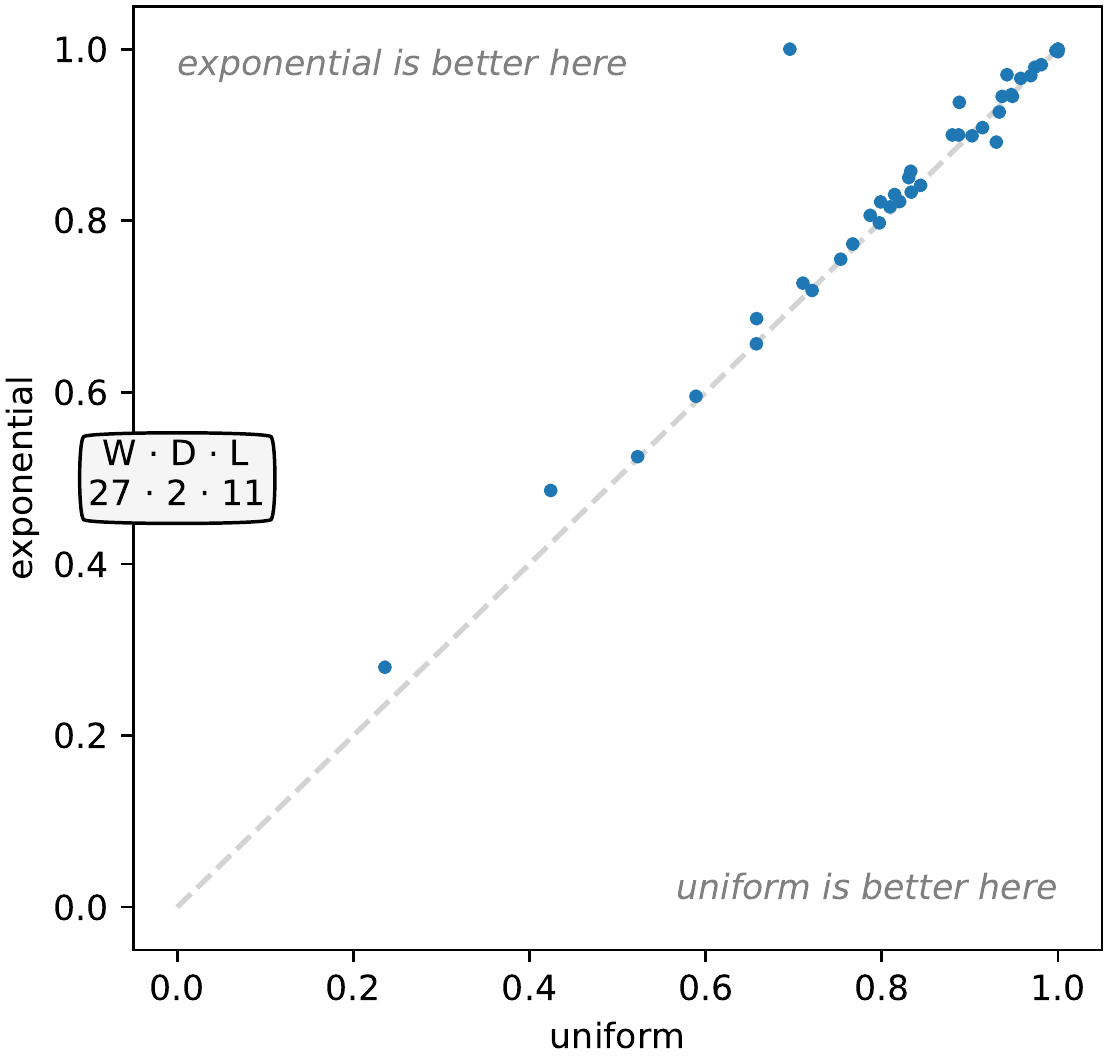}
\vspace*{-1mm}
\caption{Relative accuracy of exponential vs uniform dilation on the `development' datasets.}
\label{figure-scatter-ablation-dilation}
\end{figure}

\begin{figure}[h]
\centering
\vspace*{-1mm}
\includegraphics[width=0.35\linewidth]{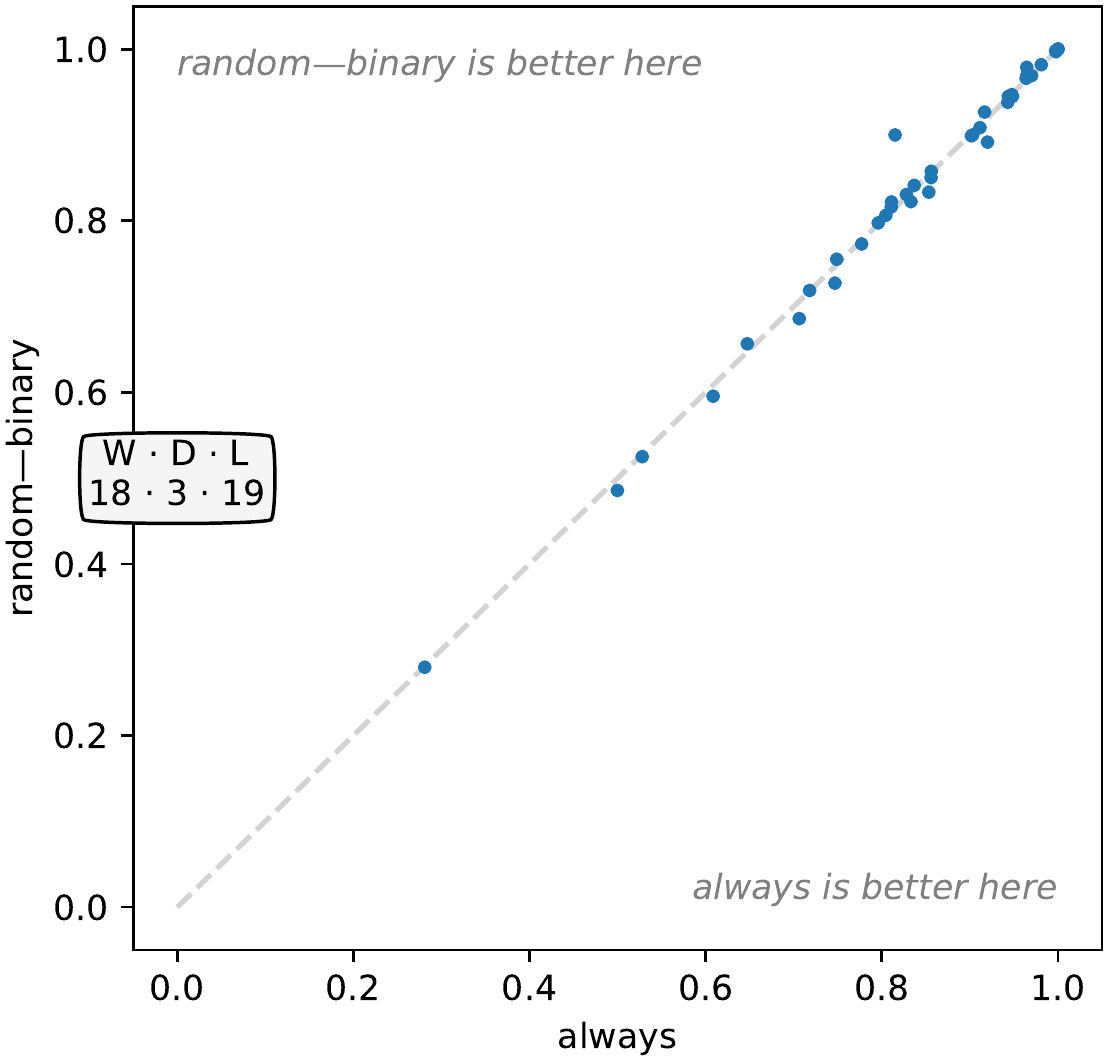}
\vspace*{-1mm}
\caption{Relative accuracy of random versus always padding on the `development' datasets.}
\label{figure-scatter-ablation-padding}
\end{figure}

\begin{figure}[h]
\centering
\vspace*{-1mm}
\includegraphics[width=0.35\linewidth]{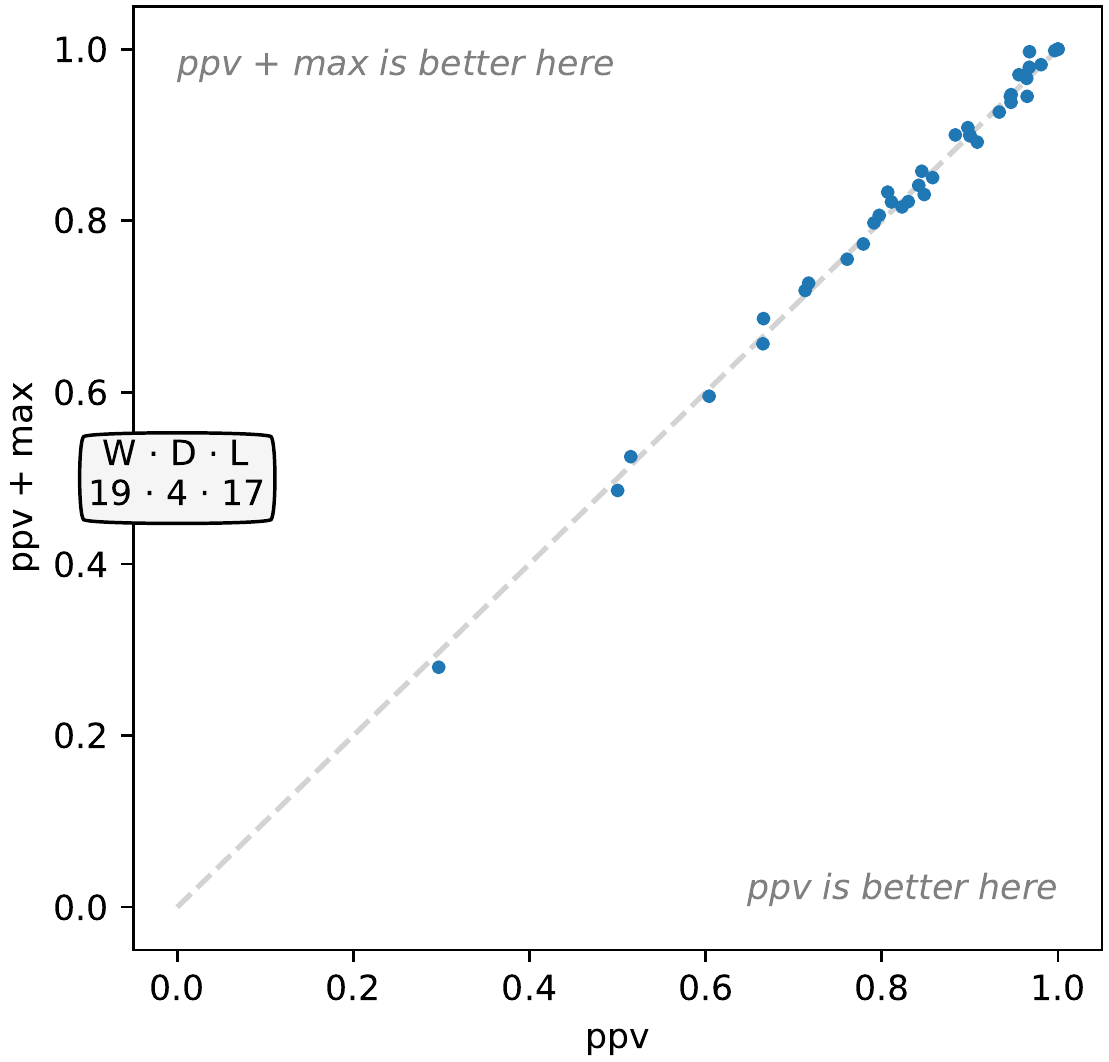}
\vspace*{-1mm}
\caption{Relative accuracy of \textit{ppv} and max versus only \textit{ppv} on the `development' datasets.}
\label{figure-scatter-ablation-features}
\end{figure}

\clearpage

\section{Results for `Bake Off' Datasets}

\textit{The `development' datasets are marked with an asterisk.}

{
\small
\tabcolsep=4pt
\begin{longtable}{lcccccccc}
\caption{Classification Accuracy, `Bake Off' Datasets}\\
\toprule
{} &  Rocket &    BOSS &      ST &    HCTE &  ResNet &      PF &   CHIEF &   ITime \\
\midrule
\endhead
\midrule
\endfoot
\bottomrule
\endlastfoot
Adiac            &  0.7847 &  0.7647 &  0.7826 &  0.8107 &  0.8289 &  0.7340 &  0.7980 &  0.8363 \\
ArrowHead        &  0.8051 &  0.8343 &  0.7371 &  0.8629 &  0.8446 &  0.8754 &  0.8229 &  0.8286 \\
*Beef            &  0.8333 &  0.8000 &  0.9000 &  0.9333 &  0.7533 &  0.7200 &  0.7333 &  0.7000 \\
BeetleFly        &  0.9000 &  0.9000 &  0.9000 &  0.9500 &  0.8500 &  0.8750 &  0.9500 &  0.8500 \\
*BirdChicken     &  0.9000 &  0.9500 &  0.8000 &  0.8500 &  0.8850 &  0.8650 &  0.9000 &  0.9500 \\
CBF              &  0.9999 &  0.9978 &  0.9744 &  0.9989 &  0.9950 &  0.9933 &  0.9978 &  0.9989 \\
*Car             &  0.8917 &  0.8333 &  0.9167 &  0.8667 &  0.9250 &  0.8467 &  0.8500 &  0.9000 \\
ChlCon           &  0.8130 &  0.6609 &  0.6997 &  0.7120 &  0.8436 &  0.6339 &  0.7206 &  0.8753 \\
CinCECGTorso     &  0.8349 &  0.8870 &  0.9543 &  0.9964 &  0.8261 &  0.9343 &  0.9826 &  0.8514 \\
Coffee           &  1.0000 &  1.0000 &  0.9643 &  1.0000 &  1.0000 &  1.0000 &  1.0000 &  1.0000 \\
Computers        &  0.7600 &  0.7560 &  0.7360 &  0.7600 &  0.8148 &  0.6444 &  0.7120 &  0.8120 \\
*CricketX        &  0.8223 &  0.7359 &  0.7718 &  0.8231 &  0.7913 &  0.8021 &  0.7974 &  0.8667 \\
*CricketY        &  0.8503 &  0.7538 &  0.7795 &  0.8487 &  0.8033 &  0.7938 &  0.8026 &  0.8513 \\
*CricketZ        &  0.8577 &  0.7462 &  0.7872 &  0.8308 &  0.8115 &  0.8010 &  0.8359 &  0.8590 \\
DiaSizRed        &  0.9703 &  0.9314 &  0.9248 &  0.9412 &  0.3013 &  0.9657 &  0.9771 &  0.9314 \\
DisPhaOutAgeGro  &  0.7547 &  0.7482 &  0.7698 &  0.7626 &  0.7165 &  0.7309 &  0.7410 &  0.7266 \\
DisPhaOutCor     &  0.7678 &  0.7283 &  0.7754 &  0.7717 &  0.7710 &  0.7928 &  0.7862 &  0.7935 \\
*DisPhaTW        &  0.7187 &  0.6763 &  0.6619 &  0.6835 &  0.6647 &  0.6597 &  0.6835 &  0.6763 \\
ECG200           &  0.9060 &  0.8700 &  0.8300 &  0.8500 &  0.8740 &  0.9090 &  0.8600 &  0.9100 \\
*ECG5000         &  0.9470 &  0.9413 &  0.9438 &  0.9462 &  0.9342 &  0.9365 &  0.9458 &  0.9409 \\
ECGFiveDays      &  1.0000 &  1.0000 &  0.9837 &  1.0000 &  0.9748 &  0.8492 &  1.0000 &  1.0000 \\
Earthquakes      &  0.7482 &  0.7482 &  0.7410 &  0.7482 &  0.7115 &  0.7540 &  0.7482 &  0.7410 \\
ElectricDevices  &  0.7305 &  0.7992 &  0.7470 &  0.7703 &  0.7291 &  0.7060 &  0.7524 &  0.7227 \\
FaceAll          &  0.9475 &  0.7817 &  0.7787 &  0.8030 &  0.8388 &  0.8938 &  0.8426 &  0.8041 \\
FaceFour         &  0.9750 &  1.0000 &  0.8523 &  0.9545 &  0.9545 &  0.9739 &  1.0000 &  0.9659 \\
FacesUCR         &  0.9616 &  0.9571 &  0.9059 &  0.9629 &  0.9547 &  0.9459 &  0.9649 &  0.9732 \\
*FiftyWords      &  0.8305 &  0.7055 &  0.7055 &  0.8088 &  0.7396 &  0.8314 &  0.8462 &  0.8418 \\
*Fish            &  0.9789 &  0.9886 &  0.9886 &  0.9886 &  0.9794 &  0.9349 &  0.9943 &  0.9829 \\
*FordA           &  0.9449 &  0.9295 &  0.9712 &  0.9644 &  0.9205 &  0.8546 &  0.9470 &  0.9483 \\
*FordB           &  0.8063 &  0.7111 &  0.8074 &  0.8235 &  0.9131 &  0.7149 &  0.8321 &  0.9365 \\
GunPoint         &  1.0000 &  1.0000 &  1.0000 &  1.0000 &  0.9907 &  0.9973 &  1.0000 &  1.0000 \\
Ham              &  0.7257 &  0.6667 &  0.6857 &  0.6667 &  0.7571 &  0.6600 &  0.7143 &  0.7143 \\
HandOutlines     &  0.9416 &  0.9027 &  0.9324 &  0.9324 &  0.9111 &  0.9214 &  0.9297 &  0.9595 \\
*Haptics         &  0.5250 &  0.4610 &  0.5227 &  0.5195 &  0.5188 &  0.4445 &  0.5162 &  0.5682 \\
*Herring         &  0.6859 &  0.5469 &  0.6719 &  0.6875 &  0.6188 &  0.5797 &  0.5781 &  0.7031 \\
InlineSkate      &  0.4582 &  0.5164 &  0.3727 &  0.5000 &  0.3731 &  0.5418 &  0.5364 &  0.4855 \\
*InsWinSou       &  0.6566 &  0.5232 &  0.6268 &  0.6551 &  0.5065 &  0.6187 &  0.6465 &  0.6348 \\
*ItaPowDem       &  0.9691 &  0.9086 &  0.9475 &  0.9631 &  0.9630 &  0.9671 &  0.9718 &  0.9679 \\
*LarKitApp       &  0.9000 &  0.7653 &  0.8587 &  0.8640 &  0.8997 &  0.7819 &  0.7893 &  0.9067 \\
Lightning2       &  0.7639 &  0.8361 &  0.7377 &  0.8197 &  0.7705 &  0.8656 &  0.7705 &  0.8033 \\
*Lightning7      &  0.8219 &  0.6849 &  0.7260 &  0.7397 &  0.8452 &  0.8219 &  0.7534 &  0.8082 \\
Mallat           &  0.9560 &  0.9382 &  0.9642 &  0.9620 &  0.9716 &  0.9576 &  0.9774 &  0.9629 \\
*Meat            &  0.9450 &  0.9000 &  0.8500 &  0.9333 &  0.9683 &  0.9333 &  0.9000 &  0.9500 \\
*MedicalImages   &  0.7975 &  0.7184 &  0.6697 &  0.7776 &  0.7703 &  0.7582 &  0.7974 &  0.7987 \\
*MidPhaOutAgeGro &  0.5955 &  0.5455 &  0.6429 &  0.5974 &  0.5688 &  0.5623 &  0.5909 &  0.5325 \\
*MidPhaOutCor    &  0.8412 &  0.7801 &  0.7938 &  0.8316 &  0.8089 &  0.8364 &  0.8522 &  0.8351 \\
MiddlePhalanxTW  &  0.5558 &  0.5455 &  0.5195 &  0.5714 &  0.4844 &  0.5292 &  0.5584 &  0.5130 \\
MoteStrain       &  0.9142 &  0.8786 &  0.8970 &  0.9329 &  0.9276 &  0.9024 &  0.9441 &  0.9034 \\
NonInvFetECGTho1 &  0.9514 &  0.8382 &  0.9496 &  0.9303 &  0.9454 &  0.9066 &  0.9074 &  0.9623 \\
NonInvFetECGTho2 &  0.9688 &  0.9008 &  0.9511 &  0.9445 &  0.9461 &  0.9399 &  0.9445 &  0.9674 \\
*OSULeaf         &  0.9380 &  0.9545 &  0.9669 &  0.9793 &  0.9785 &  0.8273 &  0.9876 &  0.9339 \\
*OliveOil        &  0.9267 &  0.8667 &  0.9000 &  0.9000 &  0.8300 &  0.8667 &  0.9000 &  0.8667 \\
PhaOutCor        &  0.8300 &  0.7716 &  0.7634 &  0.8065 &  0.8390 &  0.8235 &  0.8485 &  0.8543 \\
*Phoneme         &  0.2796 &  0.2648 &  0.3207 &  0.3824 &  0.3343 &  0.3201 &  0.3608 &  0.3354 \\
*Plane           &  1.0000 &  1.0000 &  1.0000 &  1.0000 &  1.0000 &  1.0000 &  1.0000 &  1.0000 \\
ProPhaOutAgeGro  &  0.8551 &  0.8341 &  0.8439 &  0.8585 &  0.8532 &  0.8463 &  0.8488 &  0.8537 \\
*ProPhaOutCor    &  0.8990 &  0.8488 &  0.8832 &  0.8797 &  0.9213 &  0.8732 &  0.8969 &  0.9313 \\
*ProPhaTW        &  0.8161 &  0.8000 &  0.8049 &  0.8146 &  0.7805 &  0.7790 &  0.8146 &  0.7756 \\
RefDev           &  0.5347 &  0.4987 &  0.5813 &  0.5573 &  0.5253 &  0.5323 &  0.5387 &  0.5093 \\
*ScreenType      &  0.4856 &  0.4640 &  0.5200 &  0.5893 &  0.6216 &  0.4552 &  0.5040 &  0.5760 \\
*ShapeletSim     &  1.0000 &  1.0000 &  0.9556 &  1.0000 &  0.7794 &  0.7761 &  1.0000 &  0.9889 \\
ShapesAll        &  0.9082 &  0.9083 &  0.8417 &  0.9050 &  0.9213 &  0.8858 &  0.9300 &  0.9250 \\
SmaKitApp        &  0.8213 &  0.7253 &  0.7920 &  0.8533 &  0.7861 &  0.7443 &  0.8160 &  0.7787 \\
SonAIBORobSur1   &  0.9241 &  0.6323 &  0.8436 &  0.7654 &  0.9581 &  0.8458 &  0.8270 &  0.8835 \\
SonAIBORobSur2   &  0.9164 &  0.8594 &  0.9339 &  0.9276 &  0.9778 &  0.8963 &  0.9286 &  0.9528 \\
StarLightCurves  &  0.9811 &  0.9778 &  0.9785 &  0.9815 &  0.9718 &  0.9813 &  0.9820 &  0.9792 \\
*Strawberry      &  0.9819 &  0.9757 &  0.9622 &  0.9703 &  0.9805 &  0.9684 &  0.9676 &  0.9838 \\
*SwedishLeaf     &  0.9659 &  0.9216 &  0.9280 &  0.9536 &  0.9563 &  0.9466 &  0.9664 &  0.9712 \\
Symbols          &  0.9746 &  0.9668 &  0.8824 &  0.9739 &  0.9064 &  0.9616 &  0.9799 &  0.9819 \\
*SynCon          &  0.9970 &  0.9667 &  0.9833 &  0.9967 &  0.9983 &  0.9953 &  1.0000 &  0.9967 \\
*ToeSeg1         &  0.9702 &  0.9386 &  0.9649 &  0.9825 &  0.9627 &  0.9246 &  0.9693 &  0.9693 \\
ToeSeg2          &  0.9262 &  0.9615 &  0.9077 &  0.9538 &  0.9062 &  0.8623 &  0.9538 &  0.9385 \\
*Trace           &  1.0000 &  1.0000 &  1.0000 &  1.0000 &  1.0000 &  1.0000 &  1.0000 &  1.0000 \\
TwoLeadECG       &  0.9991 &  0.9807 &  0.9974 &  0.9965 &  1.0000 &  0.9886 &  0.9965 &  0.9956 \\
TwoPatterns      &  1.0000 &  0.9930 &  0.9550 &  1.0000 &  0.9999 &  0.9996 &  1.0000 &  1.0000 \\
UWavGesLibAll    &  0.9757 &  0.9389 &  0.9422 &  0.9685 &  0.8595 &  0.9723 &  0.9687 &  0.9545 \\
UWavGesLibX      &  0.8546 &  0.7621 &  0.8029 &  0.8398 &  0.7805 &  0.8286 &  0.8417 &  0.8247 \\
*UWavGesLibY     &  0.7729 &  0.6851 &  0.7303 &  0.7655 &  0.6701 &  0.7615 &  0.7716 &  0.7688 \\
UWavGesLibZ      &  0.7917 &  0.6949 &  0.7485 &  0.7831 &  0.7501 &  0.7640 &  0.7797 &  0.7697 \\
*Wafer           &  0.9983 &  0.9948 &  1.0000 &  0.9994 &  0.9986 &  0.9955 &  0.9989 &  0.9987 \\
Wine             &  0.8074 &  0.7407 &  0.7963 &  0.7778 &  0.7444 &  0.5685 &  0.8889 &  0.6667 \\
*WordSynonyms    &  0.7552 &  0.6379 &  0.5705 &  0.7382 &  0.6224 &  0.7787 &  0.7868 &  0.7555 \\
*Worms           &  0.7273 &  0.5584 &  0.7403 &  0.5584 &  0.7909 &  0.7182 &  0.7922 &  0.8052 \\
WormsTwoClass    &  0.7987 &  0.8312 &  0.8312 &  0.7792 &  0.7468 &  0.7844 &  0.8182 &  0.7922 \\
*Yoga            &  0.9085 &  0.9183 &  0.8177 &  0.9177 &  0.8702 &  0.8786 &  0.8483 &  0.9057 \\
\end{longtable}
}

\clearpage

\section{Results for Additional 2018 Datasets}

{
\small
\tabcolsep=4pt
\begin{longtable}{lcc}
\caption{Classification Accuracy, Additional 2018 Datasets}\\
\toprule
{} &  Rocket &     DTW \\
\midrule
\endhead
ACSF1                    &  0.8780 &  0.6200 \\
AllGestureWiimoteX       &  0.7619 &  0.7171 \\
AllGestureWiimoteY       &  0.7617 &  0.7300 \\
AllGestureWiimoteZ       &  0.7491 &  0.6514 \\
BME                      &  1.0000 &  0.9800 \\
Chinatown                &  0.9802 &  0.9536 \\
Crop                     &  0.7502 &  0.7117 \\
DodgerLoopDay            &  0.5762 &  0.5875 \\
DodgerLoopGame           &  0.8725 &  0.9275 \\
DodgerLoopWeekend        &  0.9725 &  0.9783 \\
EOGHorizontalSignal      &  0.6409 &  0.4751 \\
EOGVerticalSignal        &  0.5423 &  0.4751 \\
EthanolLevel             &  0.5820 &  0.2820 \\
FreezerRegularTrain      &  0.9976 &  0.9070 \\
FreezerSmallTrain        &  0.9519 &  0.6758 \\
Fungi                    &  1.0000 &  0.8226 \\
GestureMidAirD1          &  0.8062 &  0.6385 \\
GestureMidAirD2          &  0.6831 &  0.6000 \\
GestureMidAirD3          &  0.5785 &  0.3769 \\
GesturePebbleZ1          &  0.9663 &  0.8256 \\
GesturePebbleZ2          &  0.8911 &  0.7785 \\
GunPointAgeSpan          &  0.9968 &  0.9652 \\
GunPointMaleVersusFemale &  0.9978 &  0.9747 \\
GunPointOldVersusYoung   &  0.9905 &  0.9651 \\
HouseTwenty              &  0.9639 &  0.9412 \\
InsectEPGRegularTrain    &  0.9996 &  0.8273 \\
InsectEPGSmallTrain      &  0.9815 &  0.6948 \\
MelbournePedestrian      &  0.9035 &  0.8482 \\
MixedShapesRegularTrain  &  0.9704 &  0.9089 \\
MixedShapesSmallTrain    &  0.9386 &  0.8326 \\
PLAID                    &  0.8896 &  0.8361 \\
PickupGestureWiimoteZ    &  0.8100 &  0.6600 \\
PigAirwayPressure        &  0.0885 &  0.0962 \\
PigArtPressure           &  0.9529 &  0.1971 \\
PigCVP                   &  0.9327 &  0.1587 \\
PowerCons                &  0.9311 &  0.9222 \\
Rock                     &  0.8980 &  0.8400 \\
SemgHandGenderCh2        &  0.9230 &  0.8450 \\
SemgHandMovementCh2      &  0.6444 &  0.6378 \\
SemgHandSubjectCh2       &  0.8836 &  0.8000 \\
ShakeGestureWiimoteZ     &  0.8920 &  0.8400 \\
SmoothSubspace           &  0.9793 &  0.9467 \\
UMD                      &  0.9924 &  0.9722 \\
\bottomrule
\end{longtable}
}

\end{document}